\documentclass[sigconf]{acmart}
\usepackage{booktabs}
\usepackage{multirow} 
\usepackage{tabularx} 
\usepackage{array} 
\usepackage{balance}

\AtBeginDocument{%
  }

\setcopyright{acmlicensed}
\copyrightyear{2025}
\acmYear{2025}
\acmDOI{XXXXXXX.XXXXXXX}
\acmConference[MM '25]{the 33rd ACM International Conference on Multimedia}{October 27--31, 2025}{Dublin, Ireland}
\begin{document}

\title{DIME-Net: A Dual-Illumination Adaptive Enhancement Network Based on Retinex and Mixture-of-Experts}

\author{Ziang Wang}
\affiliation{%
  \institution{Institute of Microelectronics of the Chinese Academy of Sciences}
  \city{Beijing}
  \country{China}
}
\email{wangziang24@ime.ac.cn}

\author{Xiaoqin Wang}
\authornote{Co-corresponding Author.}
\affiliation{%
  \institution{Institute of Microelectronics of the Chinese Academy of Sciences}
  \city{Beijing}
  \country{China}}
\email{wangxiaoqin@ime.ac.cn}

\author{Dingyi Wang}
\affiliation{%
  \institution{Institute of Microelectronics of the Chinese Academy of Sciences}
  \city{Beijing}
  \country{China}}
\email{wangdingyi@ime.ac.cn}

\author{Qiang Li}
\affiliation{%
  \institution{Institute of Microelectronics of the Chinese Academy of Sciences}
  \city{Beijing}
  \country{China}}
\email{liqiang@ime.ac.cn}

\author{Shushan Qiao}
\authornotemark[1]  
\affiliation{%
  \institution{Institute of Microelectronics of the Chinese Academy of Sciences}
  \city{Beijing}
  \country{China}}
\email{qiaoshushan@ime.ac.cn}
\renewcommand{\shortauthors}{Ziang Wang, Xiaoqin Wang, Dingyi Wang, Qiang Li, \& Shushan Qiao}

\begin{abstract}
Image degradation caused by complex lighting conditions such as low-light and backlit scenarios is commonly encountered in real-world environments, significantly affecting image quality and downstream vision tasks. Most existing methods focus on a single type of illumination degradation and lack the ability to handle diverse lighting conditions in a unified manner. To address this issue, we propose a dual-illumination enhancement framework called DIME-Net. The core of our method is a Mixture-of-Experts illumination estimator module, where a sparse gating mechanism adaptively selects suitable S-curve expert networks based on the illumination characteristics of the input image. By integrating Retinex theory, this module effectively performs enhancement tailored to both low-light and backlit images. To further correct illumination-induced artifacts and color distortions, we design a damage restoration module equipped with Illumination-Aware Cross Attention and Sequential-State Global Attention mechanisms. In addition, we construct a hybrid illumination dataset, MixBL, by integrating existing datasets, allowing our model to achieve robust illumination adaptability through a single training process. Experimental results show that DIME-Net achieves competitive performance on both synthetic and real-world low-light and backlit datasets without any retraining. These results demonstrate its generalization ability and potential for practical multimedia applications under diverse and complex illumination conditions.

\end{abstract}

\ccsdesc[500]{Computing methodologies~Machine learning}
\ccsdesc[500]{Computing methodologies~Computer vision problems}


\keywords{Low-Light Image Enhancement; Backlit Image Enhancement; Mixture of Experts; Global Feature Modeling}


\maketitle

\begin{figure}[htbp]
\centering
\includegraphics[scale=0.33]{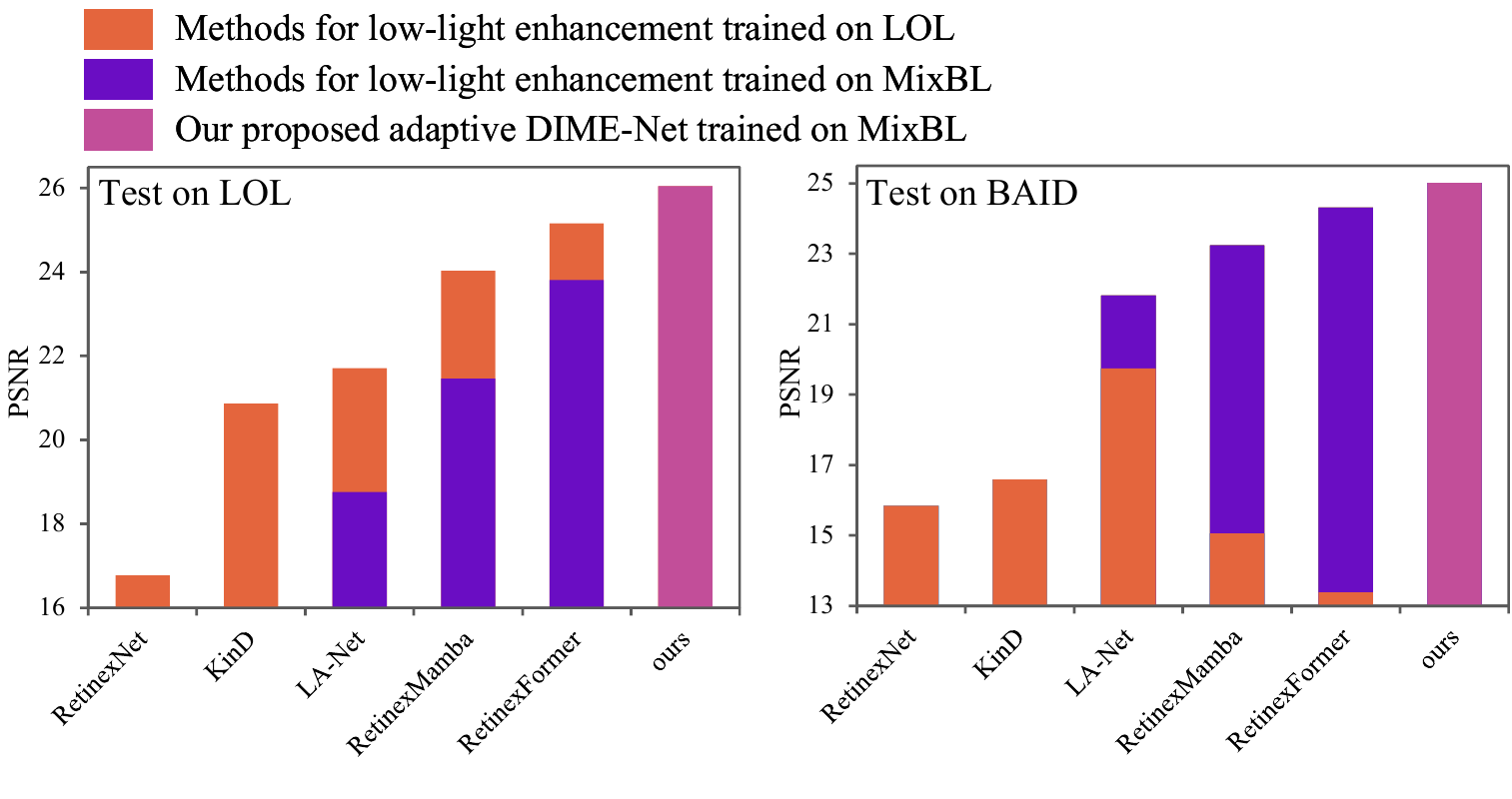}
\caption{PSNR comparison on low-light and backlit datasets. Methods trained only for low-light fail to generalize to backlit images (e.g., BAID). In contrast, our model trained on the proposed MixBL dataset achieves consistently high performance across both lighting conditions.}
\label{bar graph}
\end{figure}

\section{Introduction}

In real-world scenarios, the quality of captured images is often severely degraded due to complex ambient lighting conditions, especially in low-light or backlit environments. Such degradations typically manifest as insufficient brightness, color distortion, loss of detail, and reduced contrast. Therefore, image enhancement techniques \cite{zheng2022low,ye2024survey} have gained considerable attention for their ability to restore the perceptual quality of degraded images, making them appear more visually pleasing and closer to the human visual system’s perception in terms of brightness, color, and detail.

In recent years, a wide range of deep learning-based methods for Low-Light Image Enhancement (LLIE) \cite{lore2017llnet,li2018lightennet,lv2020fast,zhu2020eemefn,triantafyllidou2020low,li2020luminance,lu2020tbefn,lim2020dslr,chen2021improved,jin2022unsupervised,liang2022self,ye2023spatio,cai2023retinexformer,zhang2024llemamba,bai2024retinexmamba} have been proposed to restore brightness and details in images captured under poor lighting conditions. However, many existing approaches either treat Backlit Image Enhancement (BLIE) as an identical problem to LLIE, applying the same models to both, or claim to handle both degradations simultaneously \cite{zheng2022low}. As illustrated in Figure \ref {bar graph}, empirical results reveal that without retraining on backlit datasets, LLIE models often fail to enhance backlit images effectively. This observation highlights the intrinsic differences in degradation characteristics between low-light and backlit images, which are often overlooked or inadequately addressed by current enhancement methods. Although recent studies have begun to explore BLIE \cite{akaiLowArtifactFastBacklit2023,lvBacklitNetDatasetNetwork2022,liang2023iterative,gaintsevaRAVEResidualVector2024}, to the best of our knowledge, no existing method can simultaneously enhance both low-light and backlit images without compromising performance on either.

\begin{figure}[tbp]
\centering
\includegraphics[scale=0.3]{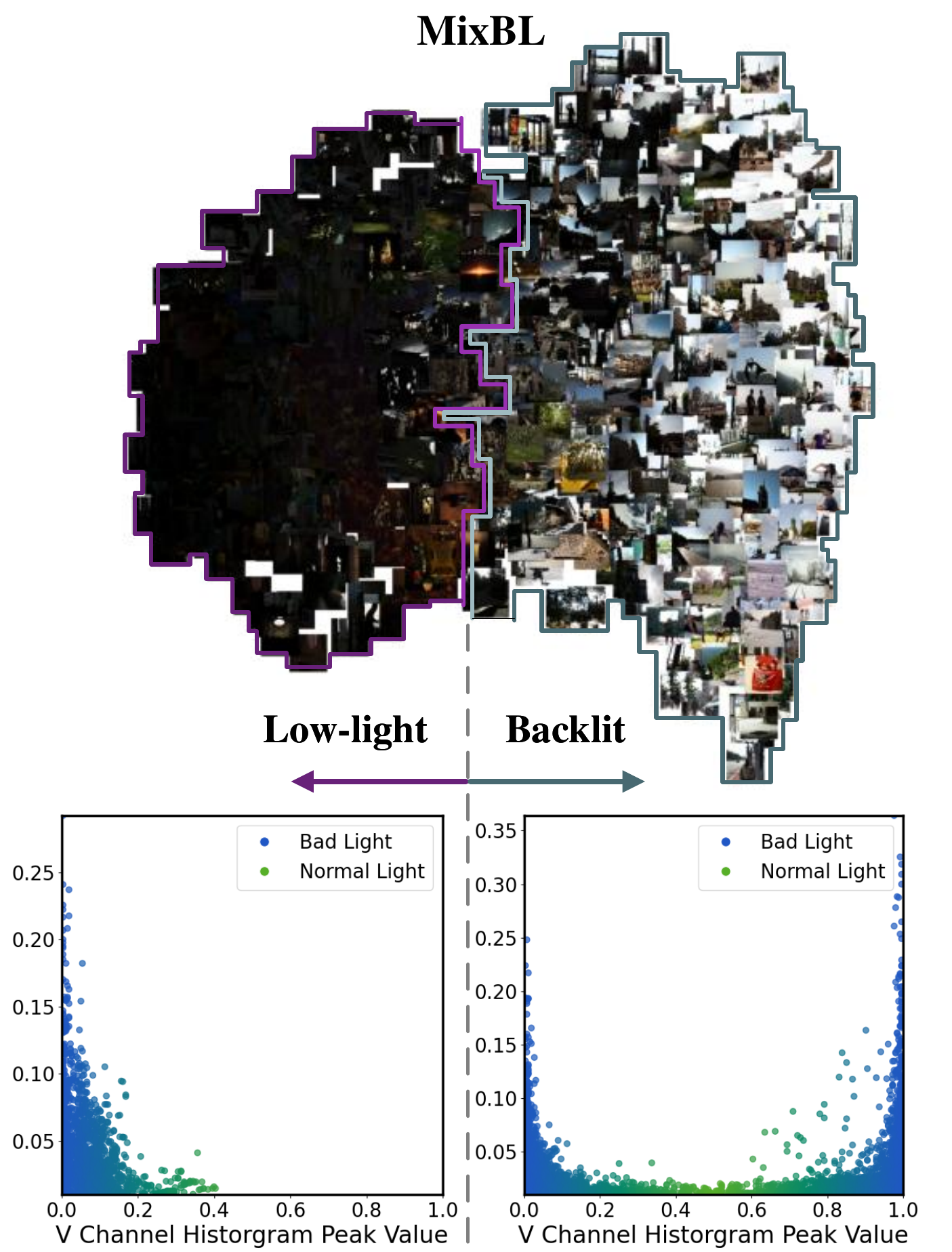}
\caption{Histogram peak distribution of brightness in the MixBL dataset. Low-light images show peaks mainly in the low-value region, while backlit images exhibit dual peaks in both low and high brightness ranges.}
\label{scatter from Mixbl_v1}
\end{figure}

To better understand the intrinsic differences, we convert images into the HSV color space \cite{smith1978color} and analyze the peak distribution of the V (brightness) channel histograms. As shown in Figure \ref {scatter from Mixbl_v1}, statistical analysis of 1,000 randomly sampled images from the LOL \cite{wei2018deep,yang2020fidelity} and BAID \cite{lvBacklitNetDatasetNetwork2022} datasets indicates that low-light images exhibit a single peak concentrated in the lower intensity range, while backlit images often show dual peaks, distributed separately in the low and high intensity ranges. This suggests a significant difference in illumination distribution: low-light degradation is primarily due to globally insufficient lighting, whereas backlit images often represent high dynamic range (HDR) \cite{yang2023learning} scenes with overexposed backgrounds and underexposed foregrounds coexisting.

Our main contributions are summarized as follows:


\begin{itemize}
  \item We propose DIME-Net, a Retinex-inspired enhancement framework consisting of an illumination estimator and a degradation restorer. The illumination estimator incorporates a sparse-gated multi-expert network, dynamically selecting enhancement strategies according to input illumination characteristics.
  \item We design a set of parameterized S-curve experts within a Mixture-of-Experts (MoE) structure, where each expert specializes in a particular illumination range. A sparse gating mechanism adaptively selects and fuses expert outputs based on input illumination, enabling enhancement across diverse lighting conditions.
  \item We build a hybrid illumination dataset (MixBL) by combining LOLv2 \cite{yang2020fidelity} and BAID \cite{lvBacklitNetDatasetNetwork2022}, and show through extensive experiments that our model trained on MixBL achieves strong generalization to both low-light and backlit datasets without retraining.

\end{itemize}

\section{Related Work}

\subsection{Low-Light Image Enhancement}
Traditional methods such as histogram equalization (HE)  
\cite{pizer1987adaptive,xiao2019histogram,abdullah2007dynamic,celik2011contextual,reza2004realization,pisano1998contrast} enhance contrast by stretching the dynamic range of pixel intensities, but both global and local HE approaches often struggle to flexibly restore fine details, resulting in limited enhancement quality. Deep learning has greatly advanced low-light enhancement by leveraging its powerful feature representation capabilities. Retinex-inspired models \cite{land1971lightness,li2018lightennet,wang2019underexposed,liu2021retinex,zhao2021retinexdip,wang2022low,ma2022toward,wu2022uretinex,cai2023retinexformer,bai2024retinexmamba}, such as Retinex-Net \cite{wei2018deep} and KinD/KinD++ \cite{zhang2019kindling,zhang2021beyond}, perform image decomposition and illumination adjustment for enhancement. Furthermore, Transformer \cite{xu2022snr,wang2023ultra,cai2023retinexformer,bai2024retinexmamba} and Mamba \cite{zouWaveMambaWaveletState2024,bai2024retinexmamba,weng2024mamballie} architectures have been introduced as global feature extractors in enhancement tasks. For example, RetinexFormer \cite{cai2023retinexformer} incorporates cross-attention mechanisms into the Retinex framework for illumination restoration, while RetinexMamba \cite{bai2024retinexmamba,zhang2024llemamba,weng2024mamballie,zou2024wave} further adopts a state-space model to enhance global feature modeling. Beyond the Retinex, network designs based on U-Net \cite{chen2018learning,guo2020zero,yang2020fidelity,fan2020integrating,xu2020learning,jiang2021enlightengan,li2021learning,zhang2022deep,dong2022abandoning,yang2023learning}, multi-branch \cite{lv2018mbllen}, multi-scale \cite{wei2018deep} structures, and nonlinear mappings are also widely used for low-light enhancement.
\subsection{Backlit Image Enhancement}
Traditional methods include segmentation-based approaches \cite{liSoftBinarySegmentationbased2015,vazquez-corralPerceptuallybasedRestorationBacklit2018,trongtirakulSingleBacklitImage2020} that divide the image into underexposed and overexposed regions for separate enhancement. For example, Akai et al. \cite{akaiSingleBacklitImage2021} used Otsu-guided filtering to construct weighting maps for enhancing bright and dark areas, and they later introduced luminance-order errors \cite{boseLumiNetMultispatialAttention2023} to further improve enhancement quality. However, these methods are computationally expensive and sensitive to scene complexity, often resulting in unstable performance. Fusion-based methods \cite{wangFusionbasedMethodSingle2016,buadesBacklitImagesEnhancement2020} enhance images using multiple tone mapping functions and fuse the results. Buades et al. \cite{buadesBacklitImagesEnhancement2020} applied gamma and logarithmic mappings and fused them via an improved Mertens algorithm to enhance contrast across RGB channels. Nevertheless, these methods often produce color distortion and unnatural illumination in extremely dark regions, lacking robustness. Deep learning approaches have been explored for backlit enhancement since 2019. ExCNet \cite{zhang2019zero} was among the first to use a block-wise loss to guide S-curve-based enhancement. BacklitNet \cite{lvBacklitNetDatasetNetwork2022} leveraged a dual-resolution structure with nested U-Net to process backlit scenes. CLIP-Lit \cite{liang2023iterative} introduced CLIP-based prompt learning to distinguish between normal and backlit images and guide the enhancement process. LumiNet \cite{boseLumiNetMultispatialAttention2023} used a modified U-Net and a conditional GAN (cGAN) to achieve more natural enhancement results.

\begin{figure*}[htbp]
\centering
\includegraphics[scale=0.15]{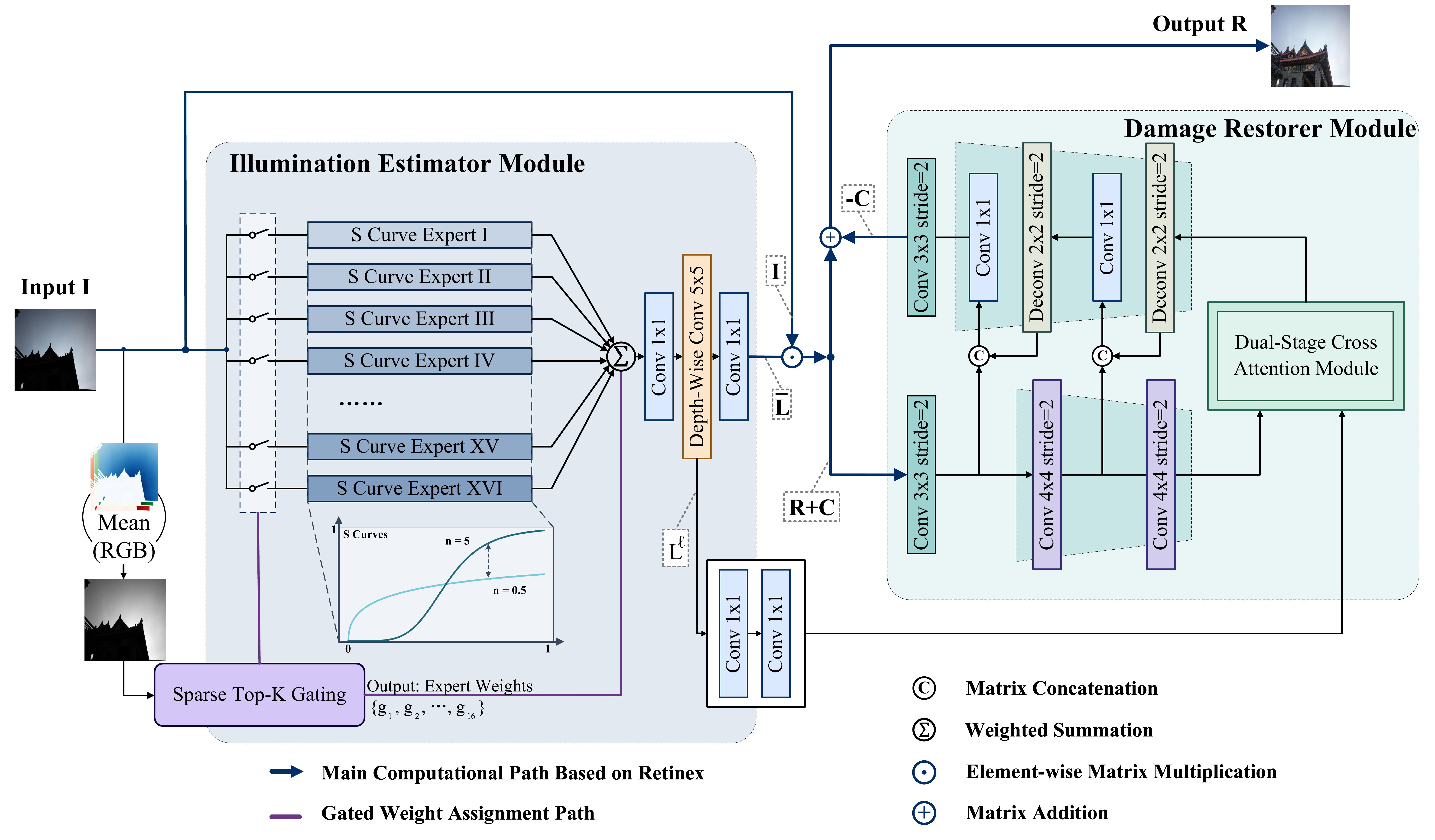}
\caption{Overview of our model architecture, primarily composed of an illumination estimator module based on a Mixture-of-Experts network and a damage restoration module employing the Dual-Stage Cross Attention Module as its bottleneck.}
\label{model_main}
\end{figure*}

\subsection{Vision Transformers vs. Vision Mamba}
Transformer models were originally introduced by Vaswani et al. \cite{vaswani2017attention} in the field of natural language processing for machine translation. Owing to their strong capability in capturing global dependencies via self-attention, Transformers have been widely adopted in high-level vision tasks such as image classification \cite{ali2021xcit,arnab2021vivit} and semantic segmentation \cite{cao2022swin,wu2021visual}. In the context of low-light enhancement, beyond RetinexFormer, which integrates cross-attention into the Retinex framework for illumination restoration, LLFormer \cite{wang2023ultra} adopts axial and cross-layer attention mechanisms to improve global information modeling.

The Mamba architecture \cite{gu2023mamba} is built upon a mathematical model known as State Space Models (SSMs) \cite{gu2022efficiently}, which have been widely used in control theory and time series analysis. Compared to Transformers, SSMs maintain long-range modeling capabilities while offering lower linear computational complexity. This architectural simplicity has made SSM-based models an attractive alternative to attention-based designs, especially for resource-constrained settings. Since the introduction of Vision Mamba \cite{xu2024visual,liu2024vmamba,pei2024efficientvmamba,huang2025enhancing}, a growing body of research has demonstrated that SSMs can serve as a general-purpose modeling tool for vision tasks. In the field of low-light enhancement, RetinexMamba was the first to incorporate SSMs into the Retinex framework, and subsequent methods such as WaveMamba \cite{zouWaveMambaWaveletState2024} and DPEC \cite{wang2024dpec} further improved performance by leveraging this structure. 

\subsection{Mixture-of-Experts (MoE) in Vision}
MoE \cite{cai2024survey} is a dynamic modeling paradigm that selectively activates a subset of expert subnetworks based on input characteristics. First introduced by Shazeer et al. \cite{shazeer2017outrageously}, sparse MoE layers use a gating network to assign weights to experts, reducing computation by only invoking selected experts. In vision, Zhu et al. \cite{zhu2022uni} proposed conditional MoE with advanced routing strategies to mitigate task and modality interference, enabling task-specific expert selection. Sun et al. \cite{cao2023multi} applied MoE to visible-infrared image fusion via a hybrid design of local (MoLE) and global (MoGE) expert modules, significantly enhancing fusion quality.

\section{Proposed Method}

The overall architecture of our proposed method, which consists of an Illumination Estimator and a Degradation Restorer, is illustrated in Figure \ref{model_main}. The illumination estimator, grounded in Retinex theory, adaptively estimates the illumination map based on the input image and a learned illumination prior using a multi-expert network composed of parameterized S-curve modules. The degradation restorer is designed to remove residual degradation from the target reflectance map and is implemented as a U-Net architecture with a Dual-Stage Cross Attention Module (DSCAM) at its bottleneck.

\subsection{Retinex-Based Adaptive Enhancement Framework}\label{AA}


According to classical Retinex theory \cite{land1971lightness}, a low-light image can be decomposed into a reflectance image $R$ and an illumination image $L$, as formulated below:
\begin{equation}
I = R \odot L
\end{equation}
where $R \in \mathbb{R}^{H \times W \times 3}$ is the reflectance map, $L \in \mathbb{R}^{H \times W \times 3}$ is the illumination map, and $\odot$ denotes element-wise multiplication. The illumination map $L$ describes the scene’s lighting intensity and distribution, while the reflectance image $R$ represents intrinsic object properties such as color, structure, and texture—essentially the appearance under ideal lighting. 

However, in real-world scenarios, the observed image $I$ is often degraded due to sensor-induced noise, artifacts, and color distortions introduced during photoelectric conversion under poor lighting. Thus, disturbance terms \cite{cai2023retinexformer,bai2024retinexmamba} are introduced into the standard Retinex model to account for such degradations:
\begin{equation}
I = \left( {R + \hat{R}} \right) \odot \left( {L + \hat{L}} \right) = R \odot L + R \odot \hat{L} + \hat{R} \odot L + \hat{R} \odot \hat{L}
\end{equation}

Specifically, $\hat{R}$ and $\hat{L}$ represent the disturbance terms in reflectance and illumination, respectively. We assume that such degradations affect both $R$ and $L$ after the image is digitized by the sensor. To address this, we introduce an illumination compensation map  $\overline{L}$ satisfying $L \odot \overline{L} = 1$, leading to the following formulation:
\begin{equation}
I \odot \overline{L} = R + R \odot \hat{L} \odot \overline{L} + \left( \hat{R} \odot \left( L + \hat{L} \right) \right) \odot \overline{L}
\label{Eq Retinex with extend}
\end{equation}
the terms $R \odot \hat{L} \odot \overline{L}$ and $\left( \hat{R} \odot \left( {L + \hat{L}} \right) \right) \odot \overline{L}$ can be collectively regarded as overall degradation C introduced to the ideally exposed image. This leads to the simplified formulation:
\begin{equation}
I \odot \overline{L} = R + C
\label{Eq Retinex with extend C}
\end{equation}

Since the illumination compensation map $\overline{L}$ is derived from the constraint $L \odot \overline{L} = 1$ it inherently encodes information about the degraded illumination. Therefore, in our approach, we design a deep learning model to jointly estimate $\overline{L}$ and the degradation term $-C$:
\begin{equation}
\left( {\overline{L},L^{l}} \right) = \mathcal{F}_{L}(I),~~~ - C = \mathcal{F}_{D}\left( {I \odot \overline{L},L^{l}} \right)
\end{equation}
we define $\mathcal{F}_L$ as the illumination estimation network that predicts $\overline{L}$ and intermediate features $L^l$ from $I$, while $\mathcal{F}_D$ estimates the degradation term $-C$. Since $C$ contains illumination-related information, $L^l$ is incorporated into the restoration:
\begin{equation}
R = {I \odot \mathcal{F}}_{L}^{(1)}(I) + \mathcal{F}_{D}\left( {{I \odot \mathcal{F}}_{L}^{(1)}(I),\mathcal{F}_{L}^{(2)}(I)} \right)
\end{equation}

The $\mathcal{F}_{L}$ is designed to learn and distinguish between the characteristics of backlit and low-light conditions, and to map these features to the corresponding illumination compensation $\overline{L}$. 

\subsection{Illumination Estimator Module Based on Mixture-of-Experts (MoE)}\label{AA}

\begin{figure}[tbp]
	\begin{minipage}{0.44\linewidth}
		\centerline{\includegraphics[width=\linewidth]{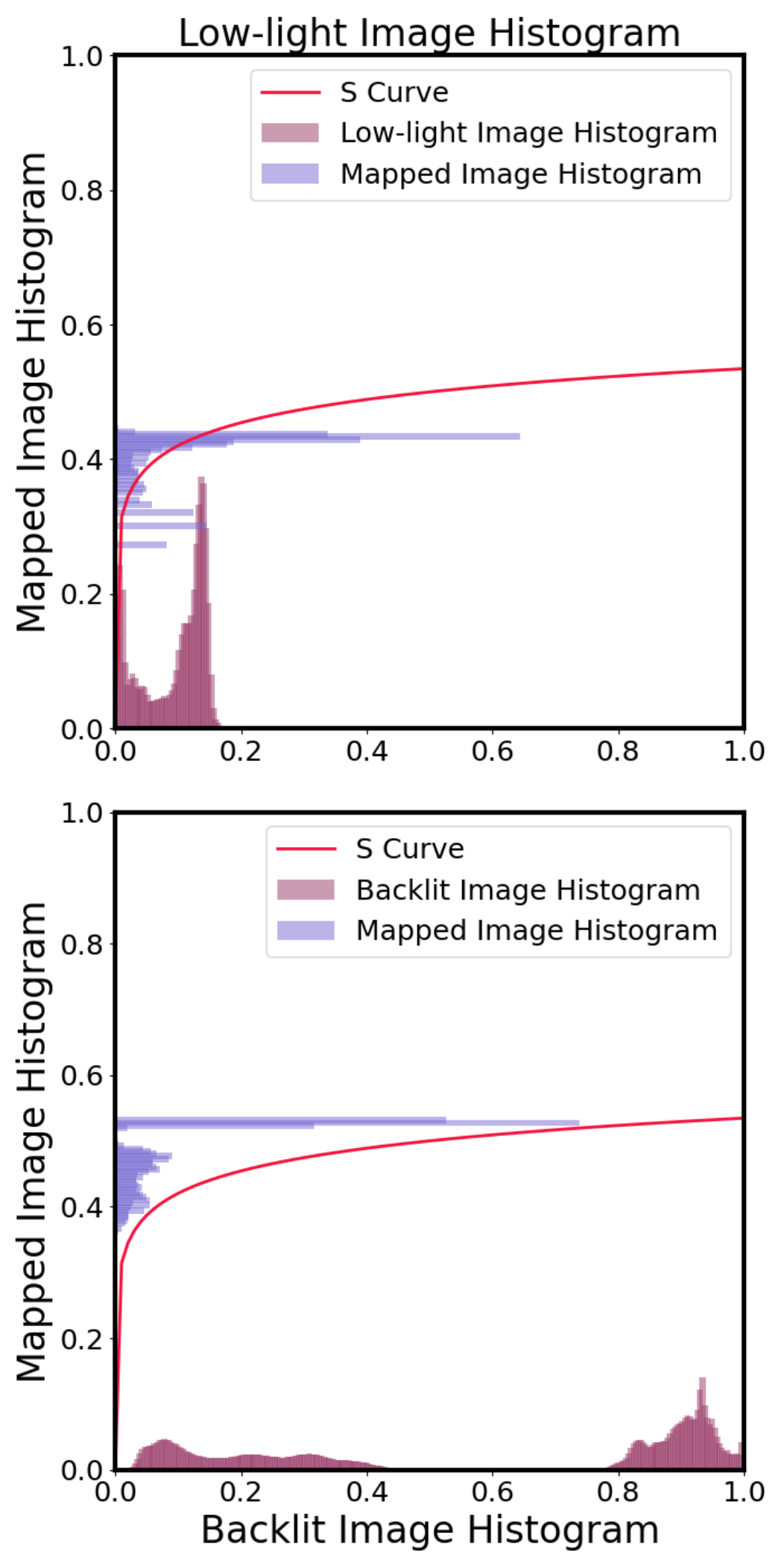}}
		\centerline{(a) S-Curve with $n = 0.2$}
	\end{minipage}
	\begin{minipage}{0.44\linewidth}
		\centerline{\includegraphics[width=\linewidth]{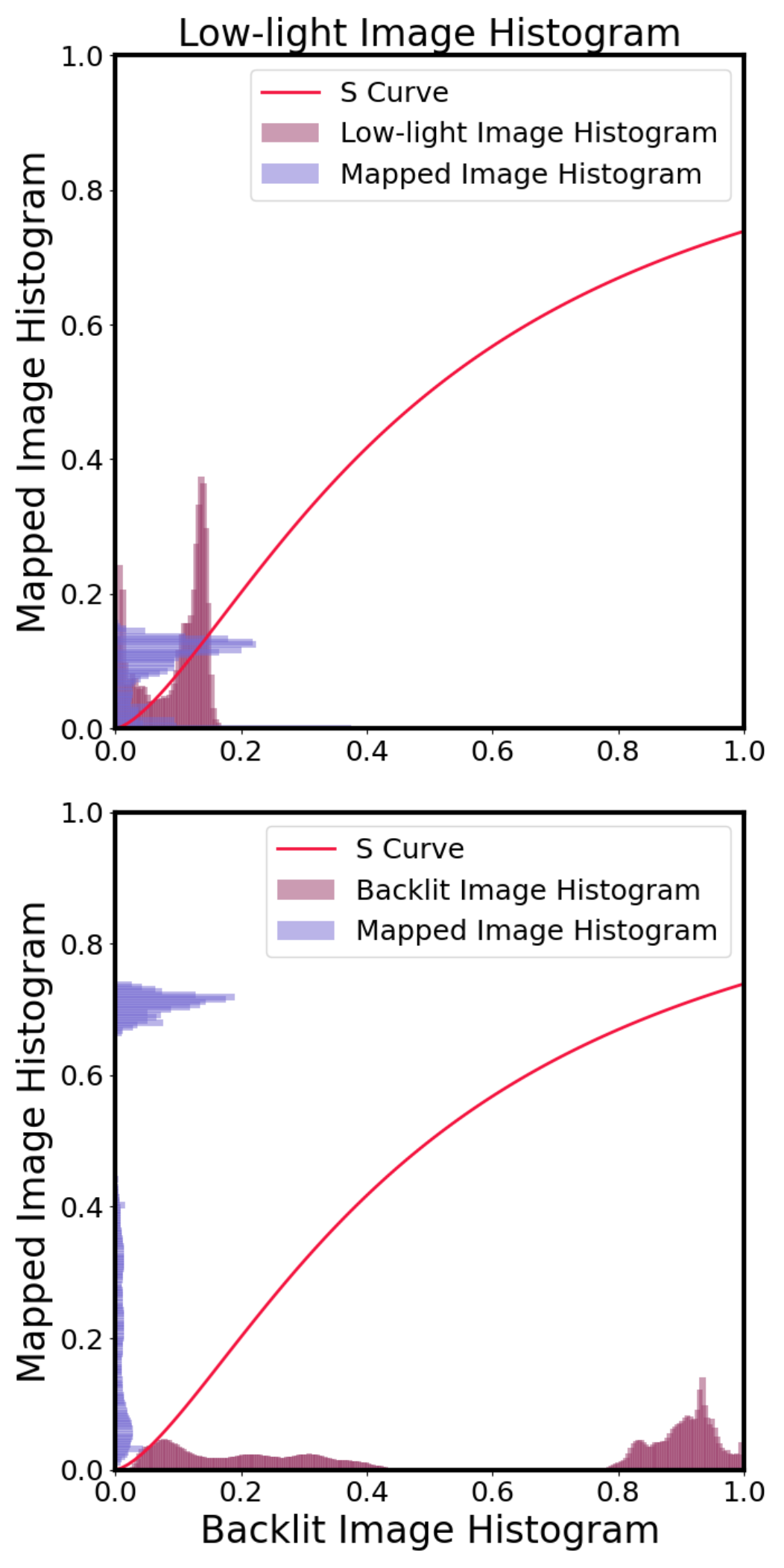}}
		\centerline{(b) S-Curve with $n = 1.5$}
	\end{minipage}
	\caption{Effect of S-Curve Parameters on Histogram Mapping for Low-Light and Backlit Images.}
	\label{S curve}
\end{figure}

As part of our illumination estimation design, we introduce a Top-$k$ sparse gating mechanism and a set of S-curve enhancement networks \cite{yang2023learning} within a MoE framework, enabling adaptive brightness compensation under varying lighting conditions. The S-curve function used in each expert is defined as follows:
\begin{equation}
f\left( {\sigma,n} \right) = \frac{I^{n}}{I^{n} + \sigma^{n}}
\end{equation}
where $I$ represents the original image intensity, and $n$ and $\sigma$ are tunable parameters that control the shape of the curve. Different values of $n$ and $\sigma$ produce distinct S-curve transformations, affecting how pixel intensities are mapped. In this work, we focus on the effect of $n$, as changes in $\sigma$ mainly shift the curve horizontally, potentially causing information loss in extremely dark or bright regions. As shown in Figure \ref{S curve}, varying $n$ leads to distinct brightness mappings: $n < 1$ enhances dark regions by pushing low intensities toward mid-tones, while $n > 1$ compresses high intensities to attenuate overexposure.


We incorporate an S-curve expert network into the Illumination Estimator module, consisting of 16 S-curve enhancement networks with different parameters n, denoted as $\left\{ {E_{f1}(I),E_{f2}(I),\ldots,E_{f16}(I)} \right\}$. Each S-curve applies a distinct nonlinear mapping to the input image I, enabling targeted intensity transformations tailored for either backlit or low-light images.

To combine the expert outputs, a sparse gating mechanism is employed, which computes gating weights based on the average RGB $I^{*}$ values of the input image. These weights are then used to aggregate the selected expert outputs, resulting in the final MoE output:
\begin{equation}
MoE(I) = \sum_{i = 1}^{n} G\!\left(I^{*}\right)_{i}\, E_{fi}(I)
\end{equation}
a larger gating weight $G_i$ indicates a stronger reliance on expert $i$, while $G_i = 0$ implies that expert $i$ is irrelevant for the current input and can be skipped during inference, thereby enabling sparse conditional computation and reducing overhead. To compute $G_i$, we follow the gating formulation introduced by Shazeer et al. \cite{shazeer2017outrageously}, which combines a deterministic projection with Gaussian noise modulation. Specifically, the gating score $H(I^*)_i$ for expert $i$ is given by:
\begin{equation}
H(I^*)_i = (I^* \cdot W_g)_i + \text{StandardNormal()} \times \text{Softplus}((I^* \cdot W_{\text{noise}})_i)
\end{equation}
where $W_g$ and $W_{\text{noise}}$ are learnable parameters, and $I^*$ is the global average RGB feature. To encourage sparsity, we retain the top-$k$ values in $H(I^*)$ and mask the rest with $-\infty$. The final gating vector is computed via softmax:
\begin{equation}
G\left( I^{*} \right) = Softmax\left( KeepTopK\left( {H\left( I^{*} \right),k} \right) \right).
\end{equation}

This sparse gating vector $G(I^*)$ weights the outputs of S-curve experts in the MoE module, enabling adaptive enhancement based on global illumination.

\subsection{Damage Restorer Module}\label{AA}

\begin{figure}[tbp]
\centering
\includegraphics[scale=0.13]{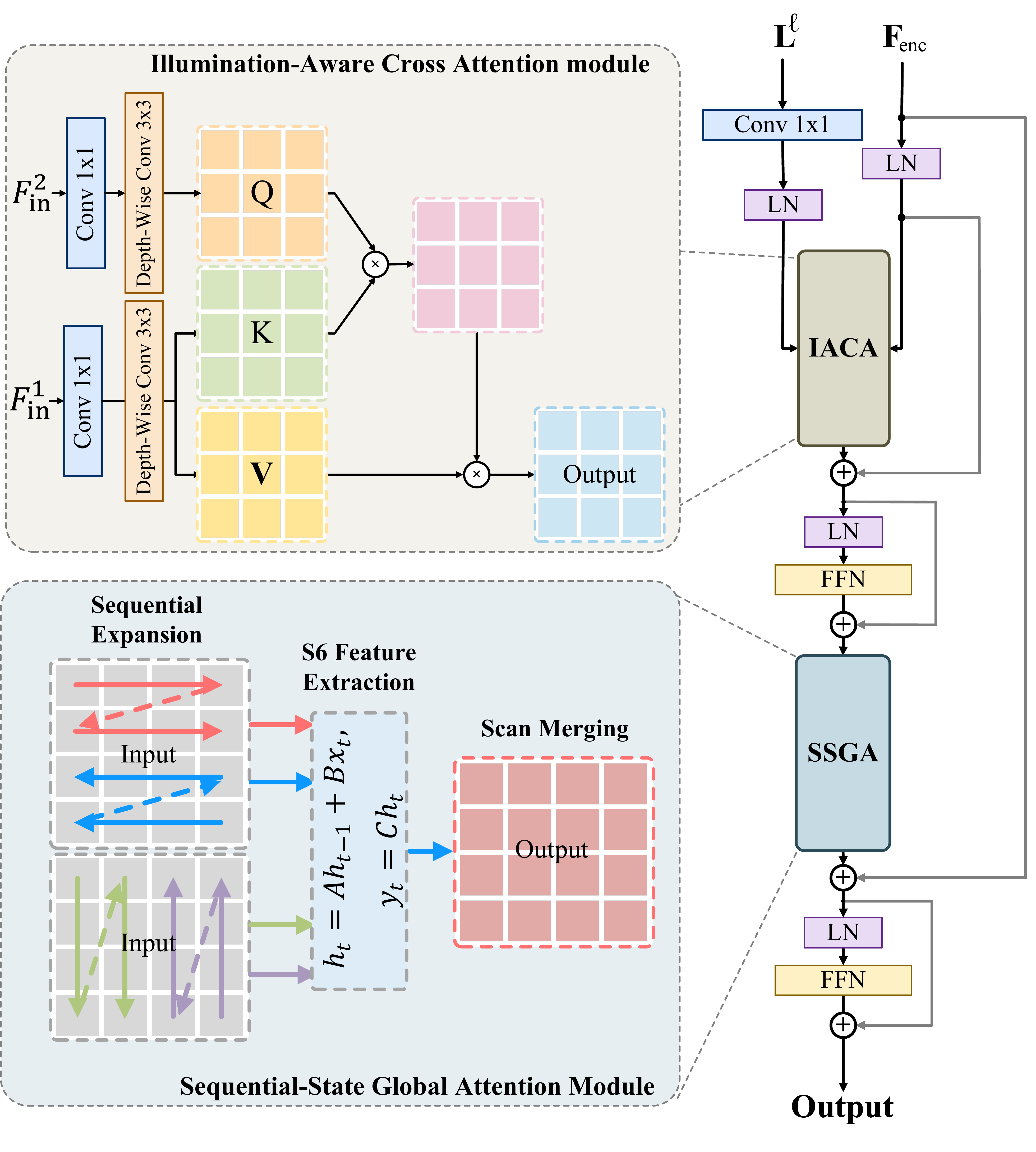}
\caption{Overview of the Damage Restorer Module. It receives encoder features $\mathbf{F}_{\text{enc}}$ and illumination features $\mathbf{L}^l$, which are processed by the IACA module for cross-attention, followed by the SSGA for global sequence modeling.}
\label{Sub_model}
\end{figure}

To address degradation issues such as noise, artifacts, and color distortion caused by poor illumination, we follow the approach of RetinexMamba \cite{bai2024retinexmamba} and adopt a U-Net architecture with a Dual-Stage Cross Attention Module (DSCAM) inserted at the bottleneck, as illustrated in the right panel of Figure \ref{model_main}. As shown in Figure \ref{Sub_model}, DSCAM is composed of two key components: Illumination-Aware Cross Attention (IACA) and Sequential-State Global Attention (SSGA).

\textbf{Illumination-Aware Cross Attention (IACA):}As shown in Equations \ref{Eq Retinex with extend} and \ref{Eq Retinex with extend C}, the degradation term $\mathbf{C}$ contains illumination-related components. IACA is designed to guide the extraction of $-\mathbf{C}$ by effectively fusing the encoder bottleneck features from the Unet ($\mathbf{F}_{\text{enc}}$) with intermediate illumination features $\mathbf{L}^l$ from the illumination estimator. To reduce complexity while preserving global illumination modeling, IACA computes attention in a channel-wise manner, treating each channel as a token. The encoder feature $\mathbf{F}_{\text{enc}} \in \mathbb{R}^{H \times W \times C}$ is first reshaped to $\mathbf{F}_{\text{in}}^1 \in \mathbb{R}^{HW \times C}$, and then split into $k$ attention heads with dimension $d_k = C / k$, resulting in $\mathbf{F}_{\text{in}}^1 \in \mathbb{R}^{HW \times d_k}$. Similarly, the illumination feature $\mathbf{L}^l$ is reshaped to $\mathbf{F}_{\text{in}}^2 \in \mathbb{R}^{HW \times d_k}$. Our method employs depthwise separable convolutions for computing the Query, Key, and Value to reduce computational complexity while enhancing local feature awareness.
\begin{equation}
K^{'},V^{'} = Conv\left( F_{in}^{1} \right),~Q^{'} = Conv\left( F_{in}^{2} \right),
\end{equation}
\begin{equation}
K = Depthwise - Conv\left( K^{'} \right),V = Depthwise - Conv\left( V^{'} \right),
\end{equation}
\begin{equation}
Q = Depthwise - Conv\left( Q^{'} \right)
\end{equation}

Finally, the attention weights are computed from the Query, Key, and Value, and used to obtain the IACA feature via weighted aggregation of the Value:
\begin{equation}
F_{IACA} = softmax\left( \frac{QK^{T}}{\alpha} \right)V
\end{equation}

This enables effective guidance from illumination features for degradation-aware restoration.

\textbf{Sequential-State Global Attention (SSGA):}Features from IACA are further refined by the SSGA module to complement matrix-based operations. SSGA employs the SS2D state-space model from the vMamba \cite{liu2024vmamba} architecture and comprises three submodules: sequential expansion, S6 feature extraction, and scan merging. Specifically, the sequential expansion module unfolds features along four directions (top-left$\leftrightarrow$bottom-right, top-right$\leftrightarrow$bottom-left) to capture multi-directional long-range dependencies. Subsequently, the S6 \cite{gu2023mamba,xu2024visual} module applies State Space Modeling (SSM) to each directional sequence $x_t$:
\begin{equation}
h_{t} = \bar{A}h_{t - 1} + \bar{B}x_{t},
\end{equation}
\begin{equation}
y_{t} = \bar{C}h_{t}
\end{equation}
where the hidden state $h_t$ encodes global illumination characteristics, and $\bar{A}$, $\bar{B}$, and $\bar{C}$ are learnable parameters, achieving linear complexity $\mathcal{O}(N)$. The S6 module extends the S4 model by integrating a selection mechanism to retain valuable features and filter redundancy. Finally, the scan merging module reconstructs features into their original spatial dimensions.

Through the synergistic integration of IACA and SSGA, the Damage Restorer module effectively suppresses noise, artifacts, and color distortions, leading to more accurate and perceptually faithful illumination restoration.


\begin{figure*}[htbp]
	
	\begin{minipage}{0.16\linewidth}
		\centerline{\includegraphics[width=\textwidth]{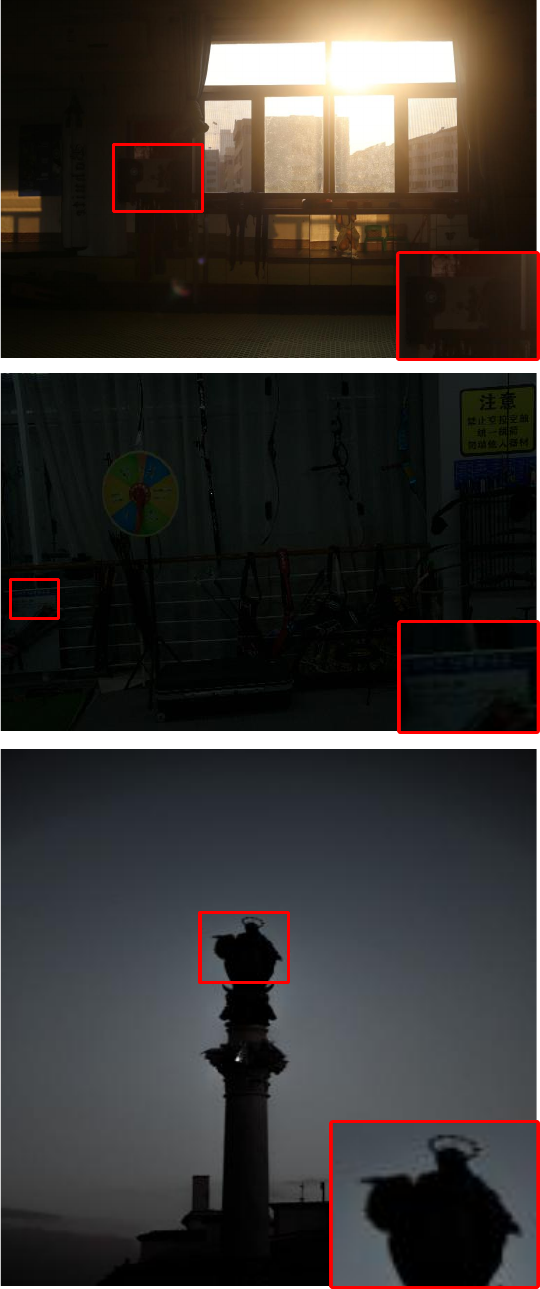}}
		\centerline{(a) Input}
	\end{minipage}
	\begin{minipage}{0.16\linewidth}
		\centerline{\includegraphics[width=\textwidth]{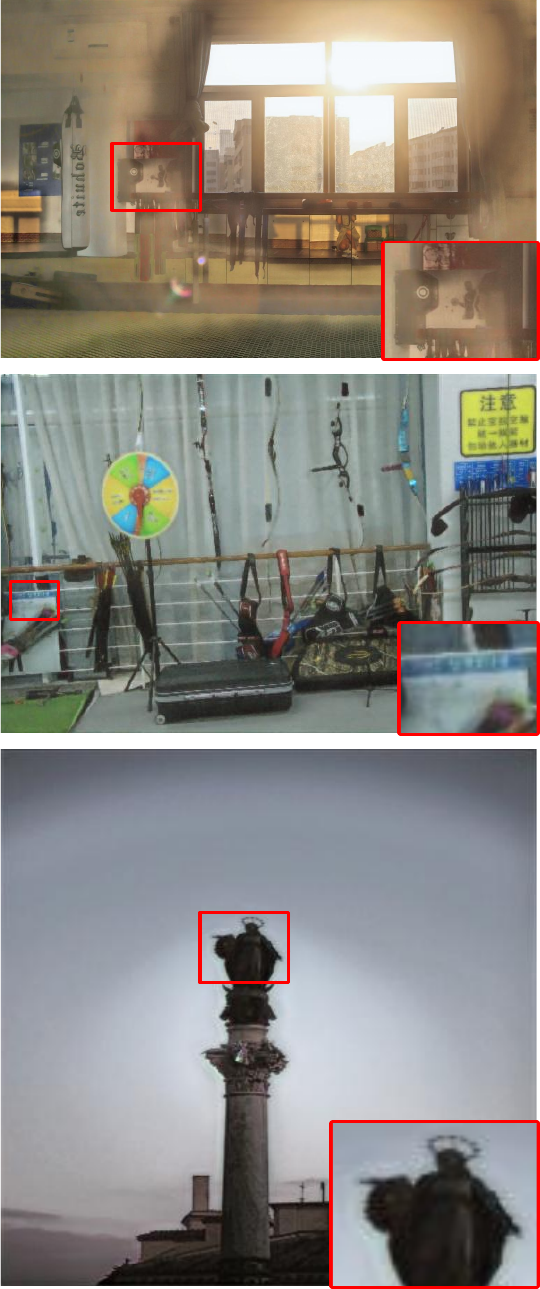}}
		\centerline{(b) LA-Net}
	\end{minipage}
	\begin{minipage}{0.16\linewidth}
		\centerline{\includegraphics[width=\textwidth]{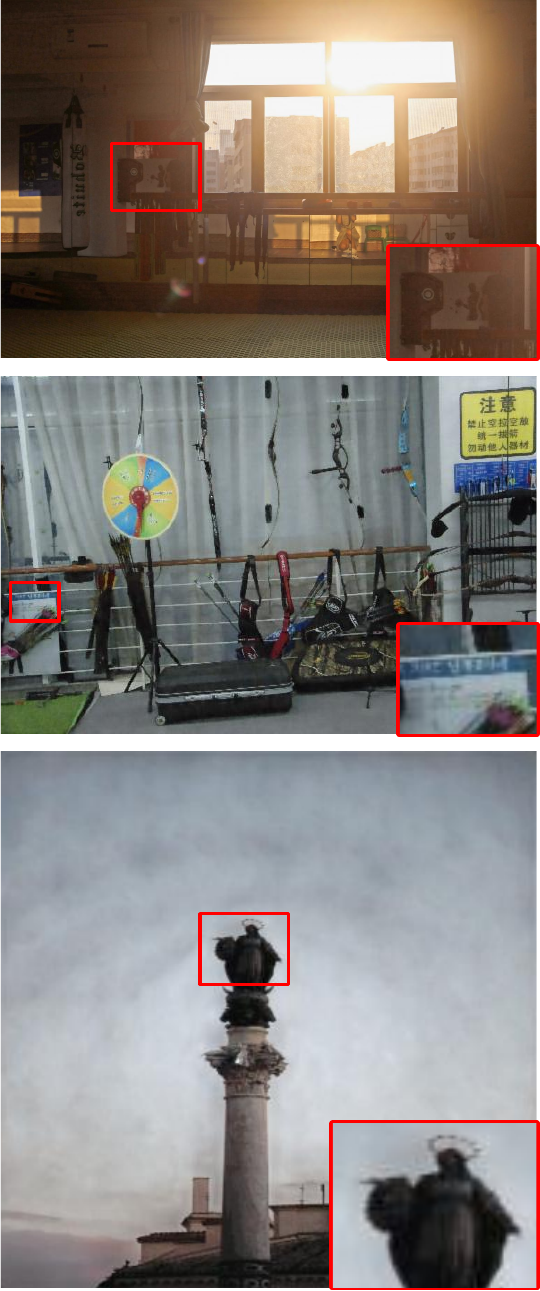}}
		\centerline{(c) RetinexMamba}
	\end{minipage}
	\begin{minipage}{0.16\linewidth}
		\centerline{\includegraphics[width=\textwidth]{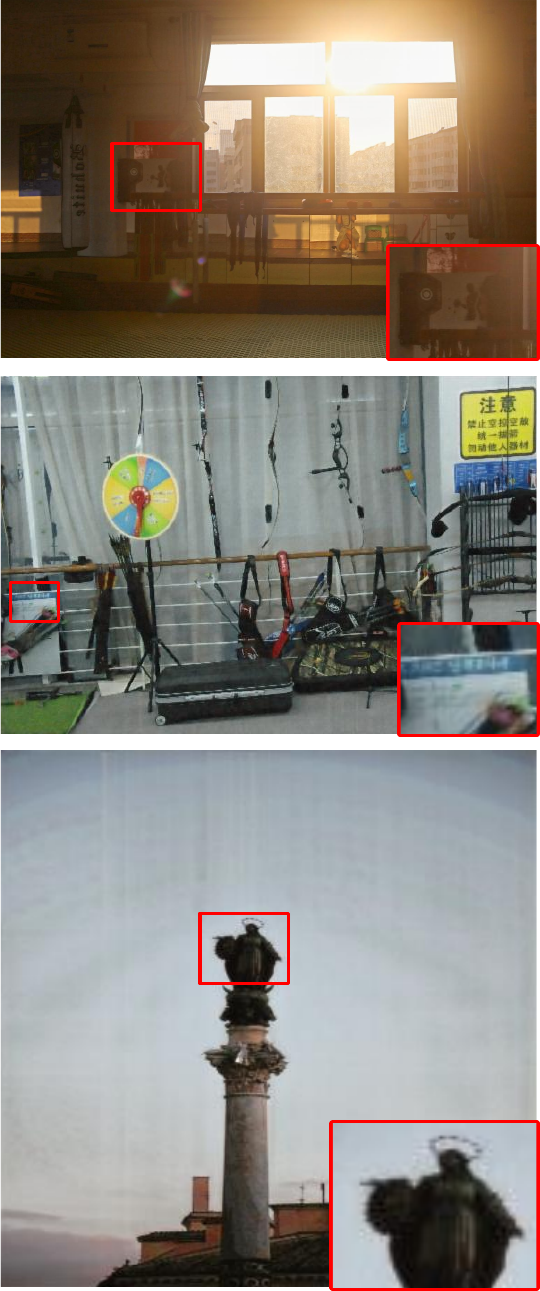}}
		\centerline{(d) RetinexFormer}
	\end{minipage}
	\begin{minipage}{0.16\linewidth}
		\centerline{\includegraphics[width=\textwidth]{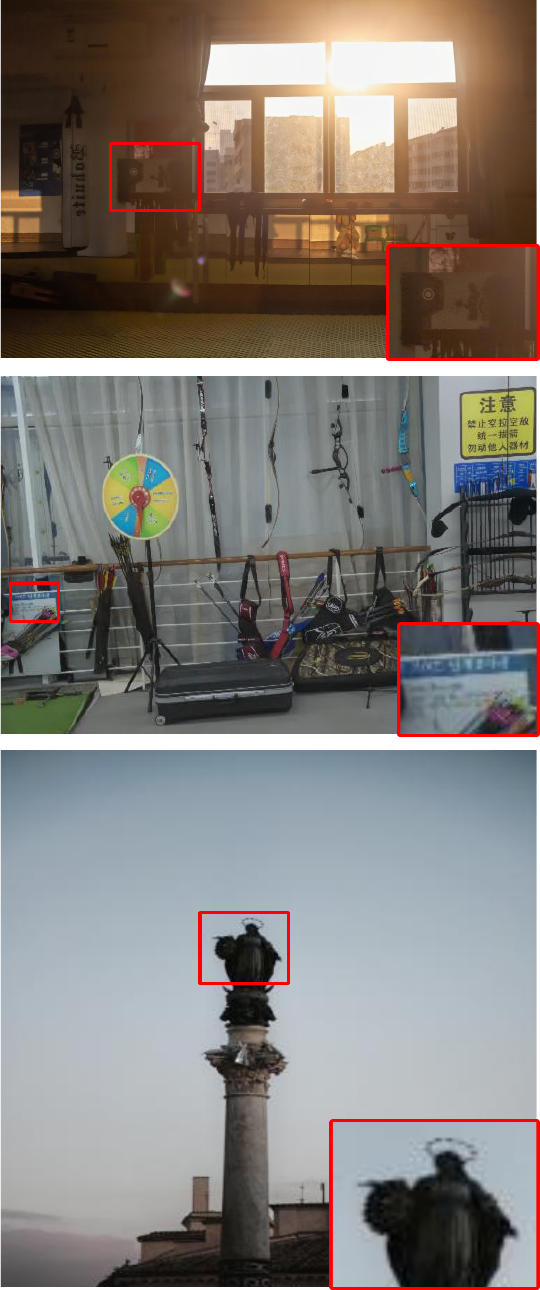}}
		\centerline{(e) Ours}
	\end{minipage}
	\begin{minipage}{0.16\linewidth}
		\centerline{\includegraphics[width=\textwidth]{Main/Q1_v1/Q1_e.pdf}}
		\centerline{(f) Ground Truth}
	\end{minipage}
 
	\caption{Qualitative comparison with baseline methods on the proposed MixBL dataset. DIME-Net effectively restores brightness and details in both low-light and backlit regions, achieving visually superior results compared to others. }
	\label{MixBL_fig}
\end{figure*}


\section{Experiments}
\subsection{Dataset}
Aiming to develop an enhancement model adaptable to both low-light and backlit conditions, we construct a mixed-illumination dataset named MixBL, comprising diverse lighting scenarios for end-to-end training. The training set includes 800 low-light images (600 real and 200 synthetic randomly selected from LOLv2) and 600 backlit images randomly selected from BAID, while the test set consists of 400 images: 100 LOLv2-real, 100 LOLv2-synthetic, and 200 BAID images, all randomly sampled. To validate the model’s adaptive enhancement capability, we evaluate the MixBL-trained model without further fine-tuning on LOLv1 \cite{wei2018deep} to assess generalization on low-light enhancement, and on BAID for backlit enhancement. We follow the original train-test splits for LOLv1 (485:15) and BAID (2500:368).


\begin{table}
\centering
\caption{Quantitative comparison on the proposed MixBL dataset, which includes both low-light and backlit images. }
\label{tab_fwdc}
\begin{tabularx}{\linewidth}{c *{3}{>{\centering\arraybackslash}X}}
\toprule[0.8pt]
\multirow{2}{*}{Methods} & \multicolumn{3}{c}{MixBL}                         \\ 
\cline{2-4}
                         & PSNR           & SSIM           & LPIPS           \\ 
\midrule[0.8pt]
LA-Net \cite{yang2023learning}                   & 12.91          & 0.784          & 0.222           \\
Retinexformer \cite{cai2023retinexformer}            & 23.12          & 0.841          & 0.146           \\
RetinexMamba \cite{bai2024retinexmamba}             & 22.45          & 0.825          & 0.170           \\ 
\midrule[0.8pt]
\textbf{DIME-Net}            & \textbf{23.98} & \textbf{0.870} & \textbf{0.100}  \\
\bottomrule[0.8pt]
\end{tabularx}
\end{table}

\subsection{Implementation}
DIME-Net is implemented in PyTorch \cite{10.5555/3454287.3455008} and trained for 600K iterations using the Adam optimizer \cite{kingma2014adam} ($\beta_1 = 0.9$, $\beta_2 = 0.999$). The initial learning rate is $2 \times 10^{-4}$ and decays to $1 \times 10^{-6}$ via cosine annealing. Training samples are $512 \times 512$ patches randomly cropped from paired low-light and backlit images in MixBL. We use a batch size of 4 and apply mixup, random rotation, and flipping for augmentation. The loss function is a weighted sum of $\ell_1$ loss, SSIM loss, and VGG perceptual loss \cite{wang2024dpec}, with respective weights of 1.0, 0.5, and 0.1. For evaluation, we adopt PSNR, SSIM \cite{wang2004image}, and LPIPS \cite{zhang2018unreasonable} as our primary evaluation metrics. For a fair comparison, all baseline models are retrained on the MixBL dataset using the training configurations specified in their respective original papers.
\begin{table*}[htbp]
\centering
\caption{Quantitative comparisons with SOTA low-light and backlit enhancement methods. LLIE-oriented methods are trained on LOLv1 and BLIE-oriented methods are trained on BAID. All baseline methods and our proposed DIME-Net are trained solely on MixBL and directly evaluated on LOLv1 and BAID without retraining or any additional dataset-specific adaptation.}
\label{LOLv1_BAID_tables}
\begin{minipage}[t]{0.49\textwidth}
\centering
\caption*{(a) Quantitative results on LOLv1 \cite{wei2018deep}}
\begin{tabularx}{\textwidth}{c c *{3}{>{\centering\arraybackslash}X}}
\toprule[0.8pt]
\multirow{2}{*}{Type} & \multirow{2}{*}{Methods} & \multicolumn{3}{c}{LOLv1} \\
\cline{3-5}
                      &                          & PSNR & SSIM & LPIPS \\
\midrule[0.8pt]
\multirow{14}{*}{\begin{tabular}[c]{@{}c@{}}LLIE-\\Oriented\\\textcolor[rgb]{0.502,0.502,0.502}{(LOLv1)}\end{tabular}} 
                      & DeepUPF \cite{ren2019low}                    & 14.38 & 0.446 & 0.367 \\
                      & Zero-DCE \cite{guo2020zero}                  & 14.86 & 0.682 & 0.372 \\
                      & RetinexNet \cite{wei2018deep}                & 16.77 & 0.560 & 0.446 \\
                      & EnGAN \cite{jiang2021enlightengan}           & 17.48 & 0.650 & 0.372 \\
                      & RUAS \cite{liu2021retinex}                    & 18.23 & 0.720 & 0.256 \\
                      & FIDE \cite{xu2020learning}                    & 18.27 & 0.665 & 0.181 \\
                      & URetinex-Net \cite{wu2022uretinex}            & 19.84 & 0.877 & 0.237 \\
                      & KinD \cite{zhang2019kindling}                 & 20.86 & 0.790 & 0.208 \\
                      & LA-Net \cite{yang2023learning}                & 21.71 & 0.810 & 0.149 \\
                      & Restormer \cite{zamir2022restormer}            & 22.43 & 0.823 & 0.139 \\
                      & RetinexMamba \cite{bai2024retinexmamba}         & 24.03 & 0.827 & 0.147 \\                      
                      & MIRNet \cite{zamirLearningEnrichedFeatures2020}  & 24.14 & 0.830 & 0.233 \\
                      & SNR-Net \cite{xu2022snr}                         & 24.61 & 0.842 & 0.150 \\
                      & RetinexFormer \cite{cai2023retinexformer}         & 25.16 & 0.845 & 0.145 \\
\hline
\multirow{3}{*}{\begin{tabular}[c]{@{}c@{}}Baseline\\\textcolor[rgb]{0.502,0.502,0.502}{(MixBL)}\end{tabular}} 
                      & LA-Net \cite{yang2023learning}              & 18.76 & 0.755 & 0.218 \\
                      & RetinexMamba \cite{bai2024retinexmamba}     & 21.46 & 0.803 & 0.201 \\
                      & RetinexFormer \cite{cai2023retinexformer}   & 23.81 & 0.832 & 0.144 \\                      
\midrule[0.8pt]
\begin{tabular}[c]{@{}c@{}}\textbf{Proposed}\\\textbf{\textcolor[rgb]{0.502,0.502,0.502}{(MixBL)}}\end{tabular} 
                      & \textbf{DIME-Net} & \textbf{26.05} & \textbf{0.869} & \textbf{0.0938} \\
\bottomrule[0.8pt]
\end{tabularx}
\end{minipage}
\hfill
\begin{minipage}[t]{0.49\textwidth}
\centering
\caption*{(b) Quantitative results on BAID \cite{lvBacklitNetDatasetNetwork2022}}
\begin{tabularx}{\textwidth}{c c *{3}{>{\centering\arraybackslash}X}}
\toprule[0.8pt]
\multirow{2}{*}{Type} & \multirow{2}{*}{Methods} & \multicolumn{3}{c}{BAID} \\
\cline{3-5}
                      &                          & PSNR & SSIM & LPIPS \\
\midrule[0.8pt]
\multirow{6}{*}{\begin{tabular}[c]{@{}c@{}}BLIE-Oriented\\\textcolor[rgb]{0.502,0.502,0.502}{(BAID)}\end{tabular}} 
                      & Li and Wu \cite{li2017learning}        & 17.16 & 0.820 & -     \\
                      & Buades et al     & 17.47 & 0.890 & -     \\
                      & ExCNet \cite{zhang2019zero}           & 19.31 & 0.900 & 0.168 \\
                      & RAVE \cite{gaintsevaRAVEResidualVector2024}             & 22.26 & 0.877 & 0.139 \\
                      & CLIP-LIT \cite{liang2023iterative}         & 21.93 & 0.875 & 0.159 \\
                      & BacklitNet \cite{lvBacklitNetDatasetNetwork2022}       & 25.06 & 0.960 & -     \\
\hline
\multirow{8}{*}{\begin{tabular}[c]{@{}c@{}}LLIE-Oriented\\\textcolor[rgb]{0.502,0.502,0.502}{(LOLv1)}\end{tabular}} 
                      & RUAS \cite{liu2021retinex}                    & 9.92  & 0.656 & 0.523 \\
                      & RetinexFormer \cite{cai2023retinexformer}    & 13.38 & 0.576 & 0.447 \\
                      & RetinexMamba \cite{bai2024retinexmamba}     & 15.06 & 0.688 & 0.351 \\
                      & RetinexNet \cite{wei2018deep}                & 15.84 & 0.695 & 0.282 \\
                      & KinD \cite{zhang2019kindling}                 & 16.59 & 0.703 & 0.383 \\
                      & EnGAN \cite{jiang2021enlightengan}            & 17.55 & 0.864 & 0.196 \\
                      & LA-Net \cite{yang2023learning}               & 17.77 & 0.714 & 0.286 \\
                      & Zero-DCE \cite{guo2020zero}                  & 19.74 & 0.870 & 0.183 \\ 
\hline
\multirow{3}{*}{\begin{tabular}[c]{@{}c@{}}Baseline\\\textcolor[rgb]{0.502,0.502,0.502}{(MixBL)}\end{tabular}} 
                      & LA-Net \cite{yang2023learning}               & 21.82 & 0.784 & 0.215 \\
                      & RetinexMamba \cite{bai2024retinexmamba}      & 23.24 & 0.802 & 0.171 \\                     
                      & RetinexFormer \cite{cai2023retinexformer}    & 24.32 & 0.814 & 0.160 \\
\midrule[0.8pt]
\begin{tabular}[c]{@{}c@{}}\textbf{Proposed}\\\textbf{\textcolor[rgb]{0.502,0.502,0.502}{(MixBL)}}\end{tabular} 
                      & \textbf{DIME-Net} & \textbf{25.02} & \textbf{0.821} & \textbf{0.101} \\
\bottomrule[0.8pt]
\end{tabularx}
\end{minipage}

\end{table*}


\begin{figure*}[htbp]
	
	\begin{minipage}[t]{0.12\linewidth}
		\centerline{\includegraphics[width=\textwidth]{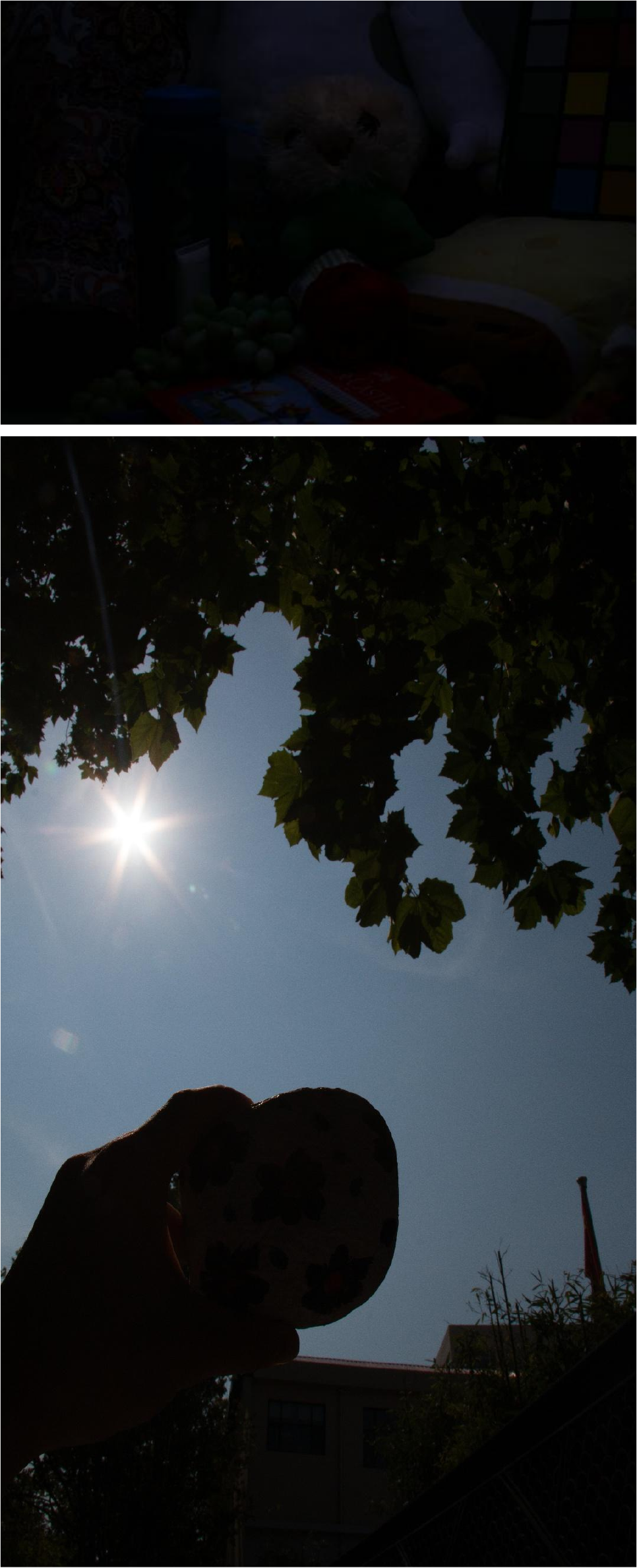}}
		\centerline{\small \shortstack{(a) Input}}
	\end{minipage}
	\begin{minipage}[t]{0.12\linewidth}		
		\centerline{\includegraphics[width=\textwidth]{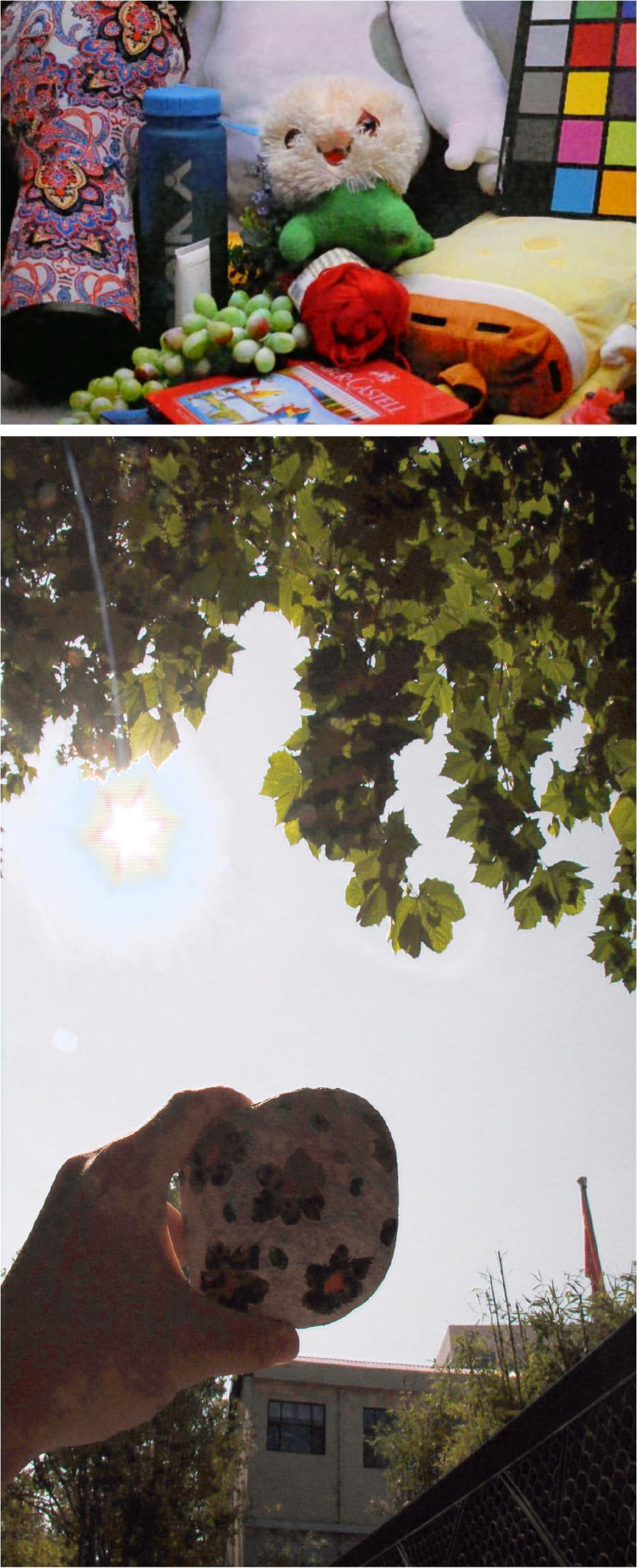}}
            \centerline{\small \shortstack{(b) RetinexFormer\\(LOLv1)}}  
	\end{minipage}
	\begin{minipage}[t]{0.12\linewidth}		
		\centerline{\includegraphics[width=\textwidth]{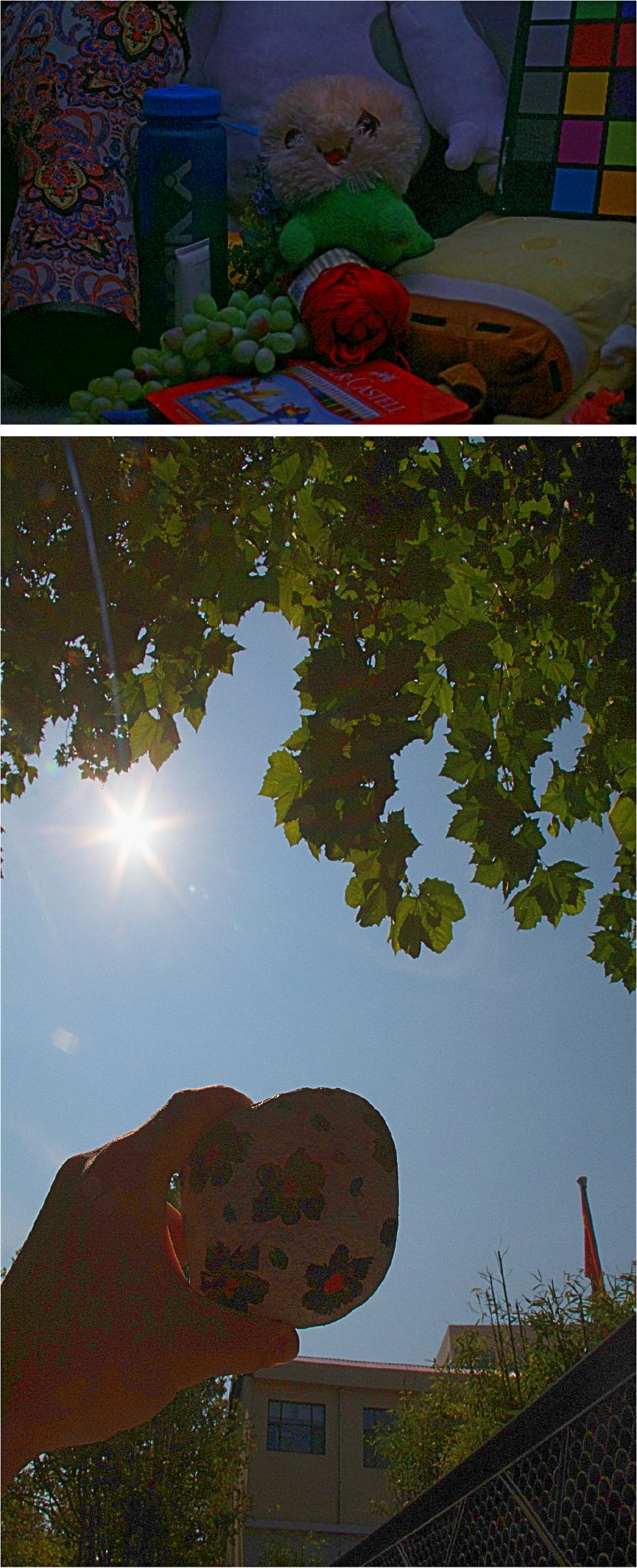}}
		\centerline{\small \shortstack{(c) CLIP-Lit\\(BAID)}}
	\end{minipage}
	\begin{minipage}[t]{0.12\linewidth}	
		\centerline{\includegraphics[width=\textwidth]{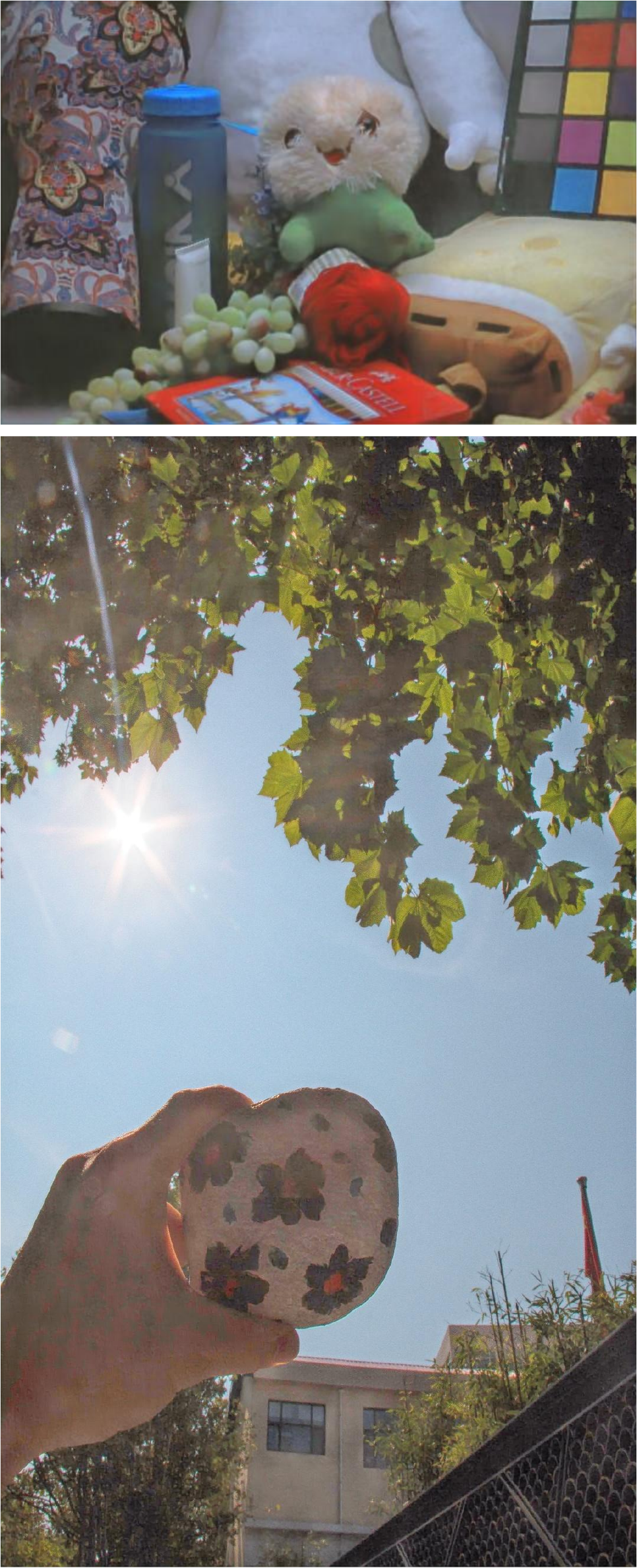}}
		\centerline{\small \shortstack{(d) LA-Net\\(MixBL)}}
	\end{minipage}
	\begin{minipage}[t]{0.12\linewidth}	
		\centerline{\includegraphics[width=\textwidth]{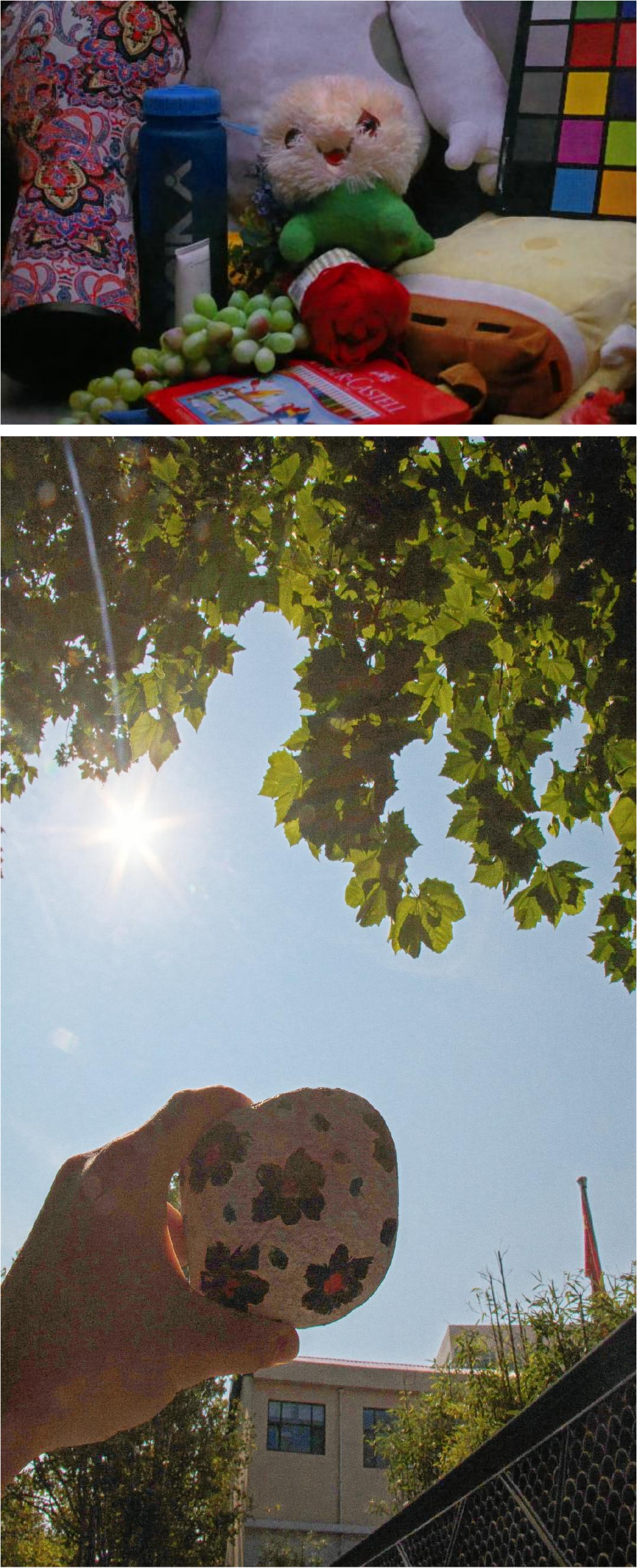}}
		\centerline{\small \shortstack{(e) RetinexFormer\\(MixBL)}}
	\end{minipage}
	\begin{minipage}[t]{0.12\linewidth}		
		\centerline{\includegraphics[width=\textwidth]{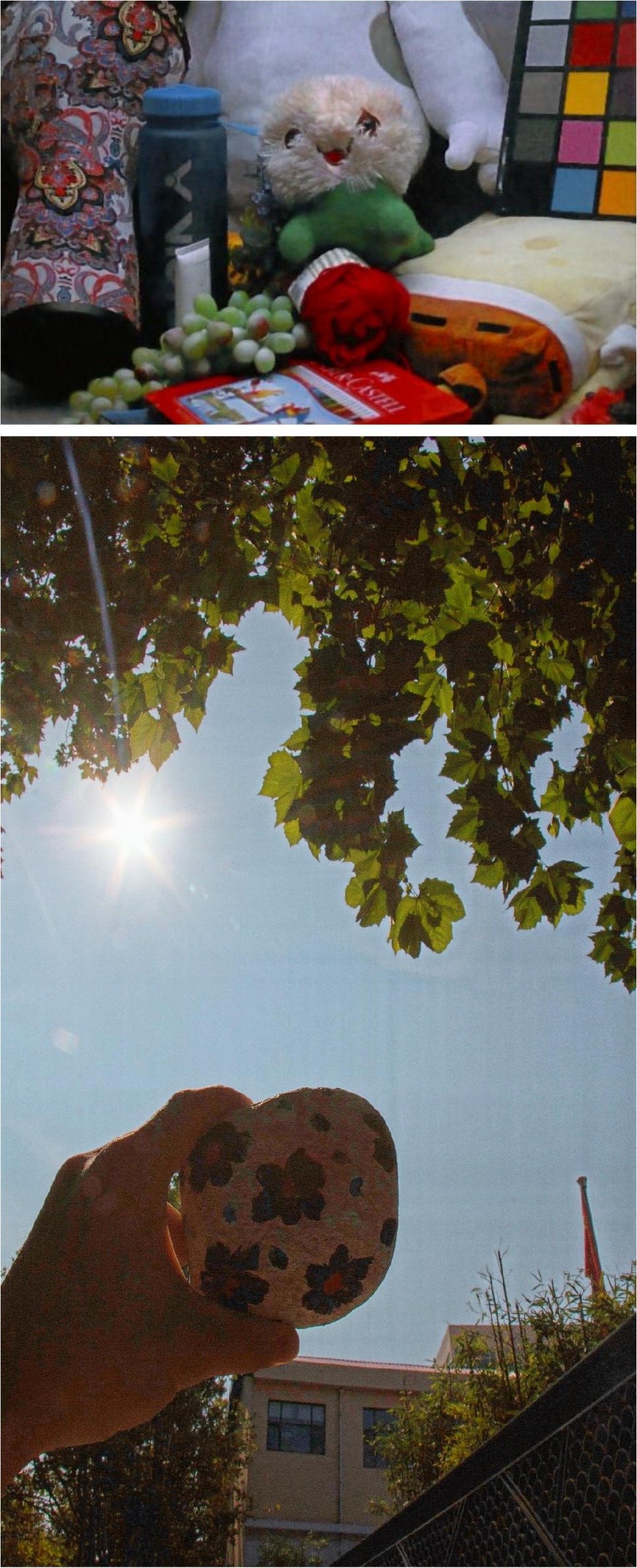}}
		\centerline{\small \shortstack{(f) RetinexMamba\\(MixBL)}}
	\end{minipage}
	\begin{minipage}[t]{0.12\linewidth}	
		\centerline{\includegraphics[width=\textwidth]{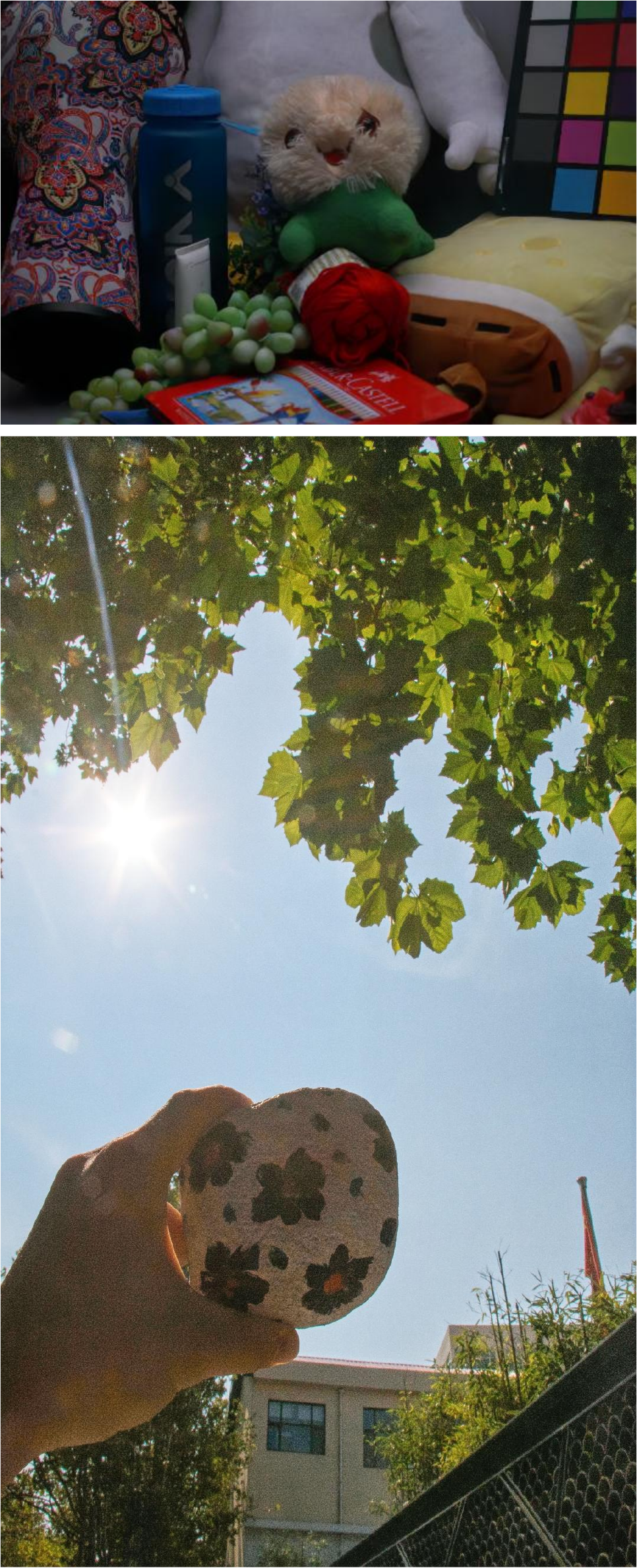}}
		\centerline{\small \shortstack{(g) Ours \\(MixBL)}}           
	\end{minipage}
	\begin{minipage}[t]{0.12\linewidth}	
		\centerline{\includegraphics[width=\textwidth]{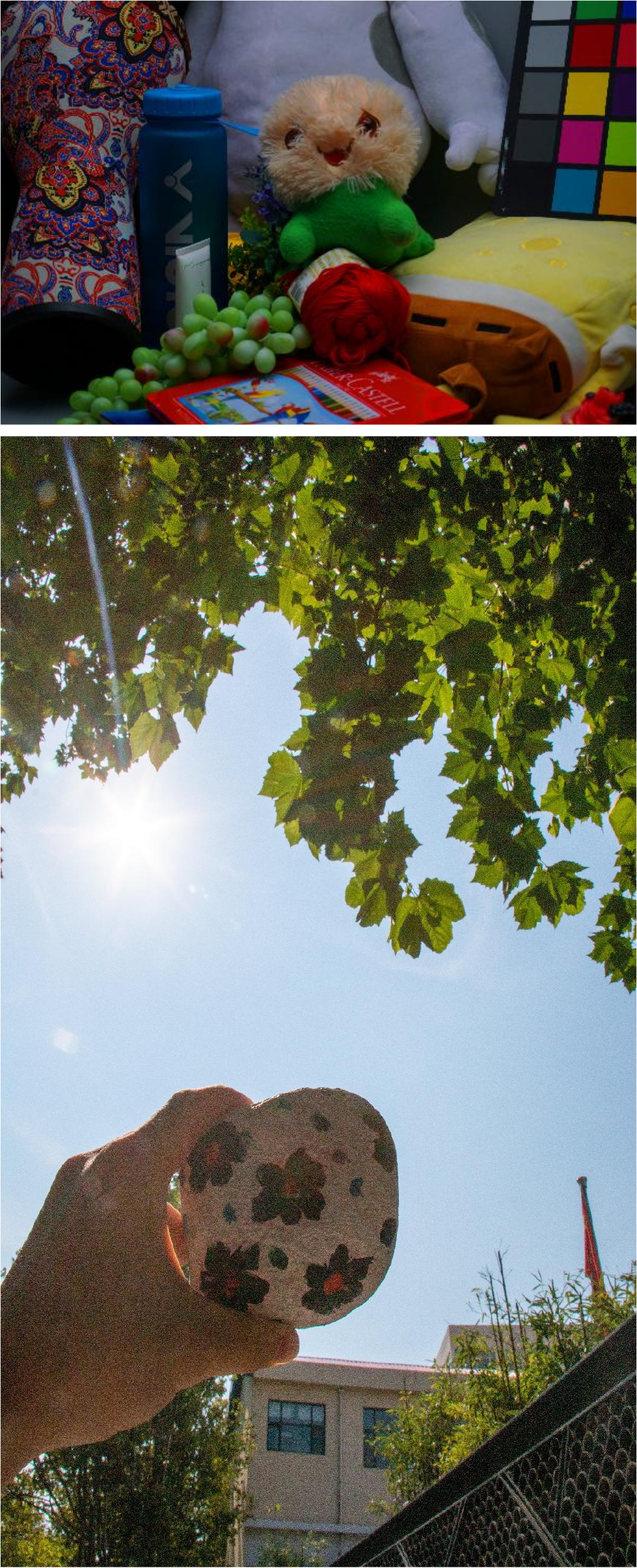}}
		\centerline{\small \shortstack{(h) Ground Truth}}
	\end{minipage}

	\caption{Qualitative comparison on the LOLv1 (top) and BAID (bottom). (b) is a low-light enhancement method trained on LOLv1, (c) is a backlit enhancement model trained on BAID. (d)–(f) are baseline methods trained on the MixBL, while (g) represents the proposed DIME-Net. The results show that DIME-Net achieves better adaptation across both degradation types. }
	\label{lolV1_BAID_fig}
\end{figure*}
 

\begin{table*}
\centering
\caption{Quantitative results of ablation study. Each variant disables or modifies a specific component of DIME-Net, which is trained solely on the MixBL dataset and directly evaluated on LOLv1 and BAID without any dataset-specific retraining.}
\label{Ablation comparison}
\begin{tabularx}{\textwidth}{c *{6}{>{\centering\arraybackslash}X}} 
\toprule[0.8pt]
\multirow{2}{*}{Experiment ~}  & \multicolumn{2}{c}{MixBL}                                               & \multicolumn{2}{c}{BAID}                                                & \multicolumn{2}{c}{LOLv1}                                                \\ 
\cline{2-7}
                           & PSNR                               & SSIM                               & PSNR                               & SSIM                               & PSNR                               & SSIM                                \\ 
\midrule[0.8pt]
w/o S-curves MoE           & 23.70                              & 0.852                              & 24.58                              & 0.818                              & 22.94                              & 0.851                               \\
All Experts with $n<1$     & 23.73                              & 0.852                              & 25.09                              & 0.821                              & 23.56                              & 0.859                               \\
All Experts with $n>1$     & 23.74                              & 0.851                              & 24.86                              & 0.819                              & 24.69                              & 0.865                               \\
w/o Illumination Estimator & 23.61                              & 0.849                              & 24.76                              & 0.817                              & 23.47                              & 0.855                               \\
w/o Damage Restorer        & 16.60                              & 0.620                              & 19.24                              & 0.747                              & 9.325                              & 0.348                               \\ 
\midrule[0.8pt]
\textbf{DIME-Net}              & \textbf{23.98} & \textbf{0.870} & \textbf{25.02} & \textbf{0.821} & \textbf{26.05} & \textbf{0.869}  \\
\bottomrule[0.8pt]
\end{tabularx}
\end{table*}

\subsection{Verification of Enhancement Capability under Mixed Illumination Conditions}

We train our adaptive enhancement model on the proposed MixBL dataset and compare it with three representative baselines: LA-Net \cite{yang2023learning}, RetinexFormer \cite{cai2023retinexformer}, and RetinexMamba \cite{bai2024retinexmamba}. LA-Net integrates S-curve enhancement into a low-frequency branch, while RetinexFormer and RetinexMamba adopt Retinex theory with degradation priors. These methods are selected to evaluate our model’s adaptability to both low-light and backlit images.

\textbf{Quantitative Analysis:} Among the selected evaluation metrics, PSNR reflects overall image fidelity and SSIM measures structural preservation—both higher values indicate better performance, while LPIPS assesses perceptual similarity, with lower values being better. Compared to the best baseline (RetinexFormer), our method improves PSNR by 0.86 dB, SSIM by 0.029, and reduces LPIPS by 0.046, indicating superior enhancement performance under mixed illumination conditions. These consistent improvements across all metrics demonstrate the robustness of our method in handling both low-light and backlit degradation.

\textbf{Qualitative Analysis:} Figure \ref{MixBL_fig} presents visual comparisons with three baselines. For backlit images, LA-Net exhibits color distortion and shadow artifacts at boundaries between bright and dark regions, while RetinexMamba and RetinexFormer fail to restore details in overexposed regions. For low-light images, LA-Net generally produces blurred results; RetinexMamba shows blurriness in magnified regions; and RetinexFormer produces overly pale colors, especially evident in the color chart region. In comparison, our method demonstrates more accurate recovery of brightness, color fidelity, and fine details for both low-light and backlit images.

\subsection{Verification of Low-Light Enhancement}

We first evaluate DIME-Net on the LOLv1 dataset, directly comparing it with other state-of-the-art (SOTA) methods specifically trained on LOLv1. Additionally, we compare our model against three baseline methods, which are also trained on MixBL without any further retraining on LOLv1.

\textbf{Quantitative Analysis:} Table \ref{LOLv1_BAID_tables}(a) presents a quantitative comparison between our method and several SOTA methods evaluated on the LOLv1 dataset. Without additional training on LOLv1, our model achieves the highest PSNR and SSIM among all state-of-the-art methods, improving PSNR by 0.89 dB over RetinexFormer. This highlights the generalization ability of our mixed-training framework for low-light enhancement tasks.

\textbf{Qualitative Analysis:} Figure \ref{lolV1_BAID_fig} shows enhancement results on the LOLv1 test set, comparing our method with RetinexFormer (LOLv1), CLIP-LIT, and three MixBL-trained baselines. Our method restores global brightness more naturally, effectively suppresses noise and artifacts, and better preserves image textures and colors, demonstrating strong adaptability to low-light conditions.

\subsection{Verification of Backlit Enhancement}

Next, we evaluate DIME-Net, trained solely on MixBL without any additional retraining, on the BAID backlit dataset to verify the effectiveness of our proposed adaptive enhancement method in backlit scenarios. We directly compare its performance against state-of-the-art (SOTA) methods specifically trained on BAID. Three baseline methods are also evaluated to make a comprehensive comparison.

\textbf{Quantitative Analysis:} Table \ref{LOLv1_BAID_tables}(b) presents a quantitative comparison of our method with other SOTA backlit enhancement methods on the BAID dataset. Our model outperforms several specialized backlit enhancement approaches, achieving up to a 3.09 dB PSNR gain over CLIP-LIT. Compared to MixBL-trained baselines, it consistently delivers better performance, confirming its ability to adaptively enhance backlit scenes by learning their degradation characteristics.

\textbf{Qualitative Analysis:} Figure \ref{lolV1_BAID_fig} presents enhancement results on the BAID dataset, comparing our method with RetinexFormer (LOLv1), CLIP-LIT, and three MixBL-trained baselines. RetinexFormer (LOLv1) produces under-enhanced results, while CLIP-LIT shows unstable colors or unnatural brightness. In contrast, our method better balances brightness and detail, delivering more natural results under backlit conditions.

\subsection{Ablation Study}

By removing or modifying specific components of our network, we train variant models on the MixBL dataset and evaluate their adaptive enhancement performance on both low-light and Backlit datasets. Specifically, we design five different architectural variants, each omitting key modules or adjusting internal parameters. The comparative results are shown in Table~\ref{Ablation comparison}.

To begin with, we analyze the effectiveness of the S-curve experts in the MoE network. When all experts adopt S-curves with $n < 1$ , the PSNR on LOLv1 drops notably (26.05 → 23.56 dB), while BAID sees a slight gain (25.02 → 25.09 dB), reflecting insufficient generalization to low-light scenes. In contrast, using only S-curves with $n > 1$ causes a performance drop on both LOLv1 and BAID, indicating limited adaptability to both low-light and backlit conditions. These observations confirm that allowing a diverse set of S-curve experts—with both $n < 1$ and $n > 1$—is essential for achieving robust dual-illumination enhancement. Furthermore, removing the entire S-curve MoE (“w/o S-curves MoE”) results in degraded performance across all datasets, validating the importance of our S-curve based MoE design and sparse gating mechanism in enabling adaptive enhancement across varying lighting conditions.

In addition, the architecture is also evaluated. Removing the Illumination Estimator module (w/o Illumination Estimator) leads to a decrease in PSNR of 1.6 dB and 0.3 dB on the LOLv1 and BAID datasets, respectively, proving the necessity of illumination priors for effective brightness correction. Finally, the experiment shows that PSNR falls to 9.325 dB and SSIM drops to 0.348 on LOLv1, while BAID also suffers a significant decrease when the damage restoration module is removed (“w/o Damage Restorer”), showing the most severe performance degradation. The results validate the critical role of our damage restoration module, which leverages cross-attention and state-space modeling to recover structural details and improve perceptual quality.

\section{Conclusion}
We introduce DIME-Net, an adaptive framework for dual-illumination enhancement that restores both low-light and backlit images. It integrates a Mixture-of-Experts illumination estimator with sparse gating to dynamically select suitable S-curve experts, guided by the Retinex theory. A degradation restoration module is designed with Illumination-Aware Cross Attention and Sequential-State Global Attention for enhanced visual quality. To support joint training, we construct MixBL, a hybrid dataset covering both lighting conditions. Experiments on LOLv1 and BAID demonstrate strong generalization, and ablation studies validate the effectiveness of each component. Future work will explore lightweight models and real-time deployment on mobile devices.


\bibliographystyle{ACM-Reference-Format}
\balance
\bibliography{DIME-Net}


\begin{thebibliography}{86}


\ifx \showCODEN    \undefined \def \showCODEN     #1{\unskip}     \fi
\ifx \showISBNx    \undefined \def \showISBNx     #1{\unskip}     \fi
\ifx \showISBNxiii \undefined \def \showISBNxiii  #1{\unskip}     \fi
\ifx \showISSN     \undefined \def \showISSN      #1{\unskip}     \fi
\ifx \showLCCN     \undefined \def \showLCCN      #1{\unskip}     \fi
\ifx \shownote     \undefined \def \shownote      #1{#1}          \fi
\ifx \showarticletitle \undefined \def \showarticletitle #1{#1}   \fi
\ifx \showURL      \undefined \def \showURL       {\relax}        \fi
\providecommand\bibfield[2]{#2}
\providecommand\bibinfo[2]{#2}
\providecommand\natexlab[1]{#1}
\providecommand\showeprint[2][]{arXiv:#2}

\bibitem[Abdullah-Al-Wadud et~al\mbox{.}(2007)]%
        {abdullah2007dynamic}
\bibfield{author}{\bibinfo{person}{Mohammad Abdullah-Al-Wadud}, \bibinfo{person}{Md~Hasanul Kabir}, \bibinfo{person}{M~Ali~Akber Dewan}, {and} \bibinfo{person}{Oksam Chae}.} \bibinfo{year}{2007}\natexlab{}.
\newblock \showarticletitle{A dynamic histogram equalization for image contrast enhancement}.
\newblock \bibinfo{journal}{\emph{IEEE transactions on consumer electronics}} \bibinfo{volume}{53}, \bibinfo{number}{2} (\bibinfo{year}{2007}), \bibinfo{pages}{593--600}.
\newblock


\bibitem[Akai et~al\mbox{.}(2021)]%
        {akaiSingleBacklitImage2021}
\bibfield{author}{\bibinfo{person}{Masato Akai}, \bibinfo{person}{Yoshiaki Ueda}, \bibinfo{person}{Takanori Koga}, {and} \bibinfo{person}{Noriaki Suetake}.} \bibinfo{year}{2021}\natexlab{}.
\newblock \showarticletitle{A {{Single Backlit Image Enhancement Method For Improvement Of Visibility Of Dark Part}}}. In \bibinfo{booktitle}{\emph{2021 {{IEEE International Conference}} on {{Image Processing}} ({{ICIP}})}}. \bibinfo{pages}{1659--1663}.
\newblock
\showISSN{2381-8549}
\href{https://doi.org/10.1109/ICIP42928.2021.9506526}{doi:\nolinkurl{10.1109/ICIP42928.2021.9506526}}


\bibitem[Akai et~al\mbox{.}(2023)]%
        {akaiLowArtifactFastBacklit2023}
\bibfield{author}{\bibinfo{person}{Masato Akai}, \bibinfo{person}{Yoshiaki Ueda}, \bibinfo{person}{Takanori Koga}, {and} \bibinfo{person}{Noriaki Suetake}.} \bibinfo{year}{2023}\natexlab{}.
\newblock \showarticletitle{Low-{{Artifact}} and {{Fast Backlit Image Enhancement Method Based}} on {{Suppression}} of {{Lightness Order Error}}}.
\newblock \bibinfo{journal}{\emph{IEEE Access}}  \bibinfo{volume}{11} (\bibinfo{year}{2023}), \bibinfo{pages}{121231--121245}.
\newblock
\showISSN{2169-3536}
\href{https://doi.org/10.1109/ACCESS.2023.3328534}{doi:\nolinkurl{10.1109/ACCESS.2023.3328534}}


\bibitem[Ali et~al\mbox{.}(2021)]%
        {ali2021xcit}
\bibfield{author}{\bibinfo{person}{Alaaeldin Ali}, \bibinfo{person}{Hugo Touvron}, \bibinfo{person}{Mathilde Caron}, \bibinfo{person}{Piotr Bojanowski}, \bibinfo{person}{Matthijs Douze}, \bibinfo{person}{Armand Joulin}, \bibinfo{person}{Ivan Laptev}, \bibinfo{person}{Natalia Neverova}, \bibinfo{person}{Gabriel Synnaeve}, \bibinfo{person}{Jakob Verbeek}, {et~al\mbox{.}}} \bibinfo{year}{2021}\natexlab{}.
\newblock \showarticletitle{Xcit: Cross-covariance image transformers}.
\newblock \bibinfo{journal}{\emph{Advances in neural information processing systems}}  \bibinfo{volume}{34} (\bibinfo{year}{2021}), \bibinfo{pages}{20014--20027}.
\newblock


\bibitem[Arnab et~al\mbox{.}(2021)]%
        {arnab2021vivit}
\bibfield{author}{\bibinfo{person}{Anurag Arnab}, \bibinfo{person}{Mostafa Dehghani}, \bibinfo{person}{Georg Heigold}, \bibinfo{person}{Chen Sun}, \bibinfo{person}{Mario Lu{\v{c}}i{\'c}}, {and} \bibinfo{person}{Cordelia Schmid}.} \bibinfo{year}{2021}\natexlab{}.
\newblock \showarticletitle{Vivit: A video vision transformer}. In \bibinfo{booktitle}{\emph{Proceedings of the IEEE/CVF international conference on computer vision}}. \bibinfo{pages}{6836--6846}.
\newblock


\bibitem[Bai et~al\mbox{.}(2024)]%
        {bai2024retinexmamba}
\bibfield{author}{\bibinfo{person}{Jiesong Bai}, \bibinfo{person}{Yuhao Yin}, \bibinfo{person}{Qiyuan He}, \bibinfo{person}{Yuanxian Li}, {and} \bibinfo{person}{Xiaofeng Zhang}.} \bibinfo{year}{2024}\natexlab{}.
\newblock \showarticletitle{Retinexmamba: Retinex-based mamba for low-light image enhancement}.
\newblock \bibinfo{journal}{\emph{arXiv preprint arXiv:2405.03349}} (\bibinfo{year}{2024}).
\newblock


\bibitem[Bose et~al\mbox{.}(2023)]%
        {boseLumiNetMultispatialAttention2023}
\bibfield{author}{\bibinfo{person}{Samprit Bose}, \bibinfo{person}{Sahil Nawale}, \bibinfo{person}{Dhruv Khut}, {and} \bibinfo{person}{Maheshkumar~H. Kolekar}.} \bibinfo{year}{2023}\natexlab{}.
\newblock \showarticletitle{{{LumiNet}}: {{Multispatial Attention Generative Adversarial Network}} for {{Backlit Image Enhancement}}}.
\newblock \bibinfo{journal}{\emph{IEEE Transactions on Instrumentation and Measurement}}  \bibinfo{volume}{72} (\bibinfo{year}{2023}), \bibinfo{pages}{1--14}.
\newblock
\showISSN{1557-9662}
\href{https://doi.org/10.1109/TIM.2023.3317384}{doi:\nolinkurl{10.1109/TIM.2023.3317384}}


\bibitem[Buades et~al\mbox{.}(2020)]%
        {buadesBacklitImagesEnhancement2020}
\bibfield{author}{\bibinfo{person}{Antoni Buades}, \bibinfo{person}{Jose-Luis Lisani}, \bibinfo{person}{Ana~Belen Petro}, {and} \bibinfo{person}{Catalina Sbert}.} \bibinfo{year}{2020}\natexlab{}.
\newblock \showarticletitle{Backlit Images Enhancement Using Global Tone Mappings and Image Fusion}.
\newblock \bibinfo{journal}{\emph{IET Image Processing}} \bibinfo{volume}{14}, \bibinfo{number}{2} (\bibinfo{year}{2020}), \bibinfo{pages}{211--219}.
\newblock
\showISSN{1751-9667}
\href{https://doi.org/10.1049/iet-ipr.2019.0814}{doi:\nolinkurl{10.1049/iet-ipr.2019.0814}}


\bibitem[Cai et~al\mbox{.}(2024)]%
        {cai2024survey}
\bibfield{author}{\bibinfo{person}{Weilin Cai}, \bibinfo{person}{Juyong Jiang}, \bibinfo{person}{Fan Wang}, \bibinfo{person}{Jing Tang}, \bibinfo{person}{Sunghun Kim}, {and} \bibinfo{person}{Jiayi Huang}.} \bibinfo{year}{2024}\natexlab{}.
\newblock \showarticletitle{A survey on mixture of experts}.
\newblock \bibinfo{journal}{\emph{arXiv preprint arXiv:2407.06204}} (\bibinfo{year}{2024}).
\newblock


\bibitem[Cai et~al\mbox{.}(2023)]%
        {cai2023retinexformer}
\bibfield{author}{\bibinfo{person}{Yuanhao Cai}, \bibinfo{person}{Hao Bian}, \bibinfo{person}{Jing Lin}, \bibinfo{person}{Haoqian Wang}, \bibinfo{person}{Radu Timofte}, {and} \bibinfo{person}{Yulun Zhang}.} \bibinfo{year}{2023}\natexlab{}.
\newblock \showarticletitle{Retinexformer: One-stage retinex-based transformer for low-light image enhancement}. In \bibinfo{booktitle}{\emph{Proceedings of the IEEE/CVF international conference on computer vision}}. \bibinfo{pages}{12504--12513}.
\newblock


\bibitem[Cao et~al\mbox{.}(2023)]%
        {cao2023multi}
\bibfield{author}{\bibinfo{person}{Bing Cao}, \bibinfo{person}{Yiming Sun}, \bibinfo{person}{Pengfei Zhu}, {and} \bibinfo{person}{Qinghua Hu}.} \bibinfo{year}{2023}\natexlab{}.
\newblock \showarticletitle{Multi-modal gated mixture of local-to-global experts for dynamic image fusion}. In \bibinfo{booktitle}{\emph{Proceedings of the IEEE/CVF international conference on computer vision}}. \bibinfo{pages}{23555--23564}.
\newblock


\bibitem[Cao et~al\mbox{.}(2022)]%
        {cao2022swin}
\bibfield{author}{\bibinfo{person}{Hu Cao}, \bibinfo{person}{Yueyue Wang}, \bibinfo{person}{Joy Chen}, \bibinfo{person}{Dongsheng Jiang}, \bibinfo{person}{Xiaopeng Zhang}, \bibinfo{person}{Qi Tian}, {and} \bibinfo{person}{Manning Wang}.} \bibinfo{year}{2022}\natexlab{}.
\newblock \showarticletitle{Swin-unet: Unet-like pure transformer for medical image segmentation}. In \bibinfo{booktitle}{\emph{European conference on computer vision}}. Springer, \bibinfo{pages}{205--218}.
\newblock


\bibitem[Celik and Tjahjadi(2011)]%
        {celik2011contextual}
\bibfield{author}{\bibinfo{person}{Turgay Celik} {and} \bibinfo{person}{Tardi Tjahjadi}.} \bibinfo{year}{2011}\natexlab{}.
\newblock \showarticletitle{Contextual and variational contrast enhancement}.
\newblock \bibinfo{journal}{\emph{IEEE Transactions on Image Processing}} \bibinfo{volume}{20}, \bibinfo{number}{12} (\bibinfo{year}{2011}), \bibinfo{pages}{3431--3441}.
\newblock


\bibitem[Chen et~al\mbox{.}(2018)]%
        {chen2018learning}
\bibfield{author}{\bibinfo{person}{Chen Chen}, \bibinfo{person}{Qifeng Chen}, \bibinfo{person}{Jia Xu}, {and} \bibinfo{person}{Vladlen Koltun}.} \bibinfo{year}{2018}\natexlab{}.
\newblock \showarticletitle{Learning to see in the dark}. In \bibinfo{booktitle}{\emph{Proceedings of the IEEE conference on computer vision and pattern recognition}}. \bibinfo{pages}{3291--3300}.
\newblock


\bibitem[Chen et~al\mbox{.}(2021)]%
        {chen2021improved}
\bibfield{author}{\bibinfo{person}{Qili Chen}, \bibinfo{person}{Junfang Fan}, {and} \bibinfo{person}{Wenbai Chen}.} \bibinfo{year}{2021}\natexlab{}.
\newblock \showarticletitle{An improved image enhancement framework based on multiple attention mechanism}.
\newblock \bibinfo{journal}{\emph{Displays}}  \bibinfo{volume}{70} (\bibinfo{year}{2021}), \bibinfo{pages}{102091}.
\newblock


\bibitem[Dong et~al\mbox{.}(2022)]%
        {dong2022abandoning}
\bibfield{author}{\bibinfo{person}{Xingbo Dong}, \bibinfo{person}{Wanyan Xu}, \bibinfo{person}{Zhihui Miao}, \bibinfo{person}{Lan Ma}, \bibinfo{person}{Chao Zhang}, \bibinfo{person}{Jiewen Yang}, \bibinfo{person}{Zhe Jin}, \bibinfo{person}{Andrew Beng~Jin Teoh}, {and} \bibinfo{person}{Jiajun Shen}.} \bibinfo{year}{2022}\natexlab{}.
\newblock \showarticletitle{Abandoning the bayer-filter to see in the dark}. In \bibinfo{booktitle}{\emph{Proceedings of the ieee/cvf conference on computer vision and pattern recognition}}. \bibinfo{pages}{17431--17440}.
\newblock


\bibitem[Fan et~al\mbox{.}(2020)]%
        {fan2020integrating}
\bibfield{author}{\bibinfo{person}{Minhao Fan}, \bibinfo{person}{Wenjing Wang}, \bibinfo{person}{Wenhan Yang}, {and} \bibinfo{person}{Jiaying Liu}.} \bibinfo{year}{2020}\natexlab{}.
\newblock \showarticletitle{Integrating semantic segmentation and retinex model for low-light image enhancement}. In \bibinfo{booktitle}{\emph{Proceedings of the 28th ACM international conference on multimedia}}. \bibinfo{pages}{2317--2325}.
\newblock


\bibitem[Gaintseva et~al\mbox{.}(2024)]%
        {gaintsevaRAVEResidualVector2024}
\bibfield{author}{\bibinfo{person}{Tatiana Gaintseva}, \bibinfo{person}{Martin Benning}, {and} \bibinfo{person}{Gregory Slabaugh}.} \bibinfo{year}{2024}\natexlab{}.
\newblock \bibinfo{title}{{{RAVE}}: {{Residual Vector Embedding}} for {{CLIP-Guided Backlit Image Enhancement}}}.
\newblock
\href{https://doi.org/10.48550/arXiv.2404.01889}{doi:\nolinkurl{10.48550/arXiv.2404.01889}}
\showeprint[arxiv]{2404.01889}~[cs]


\bibitem[Gu and Dao(2023)]%
        {gu2023mamba}
\bibfield{author}{\bibinfo{person}{Albert Gu} {and} \bibinfo{person}{Tri Dao}.} \bibinfo{year}{2023}\natexlab{}.
\newblock \showarticletitle{Mamba: Linear-time sequence modeling with selective state spaces}.
\newblock \bibinfo{journal}{\emph{arXiv preprint arXiv:2312.00752}} (\bibinfo{year}{2023}).
\newblock


\bibitem[Gu et~al\mbox{.}(2022)]%
        {gu2022efficiently}
\bibfield{author}{\bibinfo{person}{Albert Gu}, \bibinfo{person}{Karan Goel}, {and} \bibinfo{person}{Christopher Re}.} \bibinfo{year}{2022}\natexlab{}.
\newblock \showarticletitle{Efficiently Modeling Long Sequences with Structured State Spaces}. In \bibinfo{booktitle}{\emph{International Conference on Learning Representations}}.
\newblock
\urldef\tempurl%
\url{https://openreview.net/forum?id=uYLFoz1vlAC}
\showURL{%
\tempurl}


\bibitem[Guo et~al\mbox{.}(2020)]%
        {guo2020zero}
\bibfield{author}{\bibinfo{person}{Chunle Guo}, \bibinfo{person}{Chongyi Li}, \bibinfo{person}{Jichang Guo}, \bibinfo{person}{Chen~Change Loy}, \bibinfo{person}{Junhui Hou}, \bibinfo{person}{Sam Kwong}, {and} \bibinfo{person}{Runmin Cong}.} \bibinfo{year}{2020}\natexlab{}.
\newblock \showarticletitle{Zero-reference deep curve estimation for low-light image enhancement}. In \bibinfo{booktitle}{\emph{Proceedings of the IEEE/CVF conference on computer vision and pattern recognition}}. \bibinfo{pages}{1780--1789}.
\newblock


\bibitem[Huang et~al\mbox{.}(2025)]%
        {huang2025enhancing}
\bibfield{author}{\bibinfo{person}{Jiahao Huang}, \bibinfo{person}{Liutao Yang}, \bibinfo{person}{Fanwen Wang}, \bibinfo{person}{Yinzhe Wu}, \bibinfo{person}{Yang Nan}, \bibinfo{person}{Weiwen Wu}, \bibinfo{person}{Chengyan Wang}, \bibinfo{person}{Kuangyu Shi}, \bibinfo{person}{Angelica~I Aviles-Rivero}, \bibinfo{person}{Carola-Bibiane Sch{\"o}nlieb}, {et~al\mbox{.}}} \bibinfo{year}{2025}\natexlab{}.
\newblock \showarticletitle{Enhancing global sensitivity and uncertainty quantification in medical image reconstruction with Monte Carlo arbitrary-masked mamba}.
\newblock \bibinfo{journal}{\emph{Medical Image Analysis}}  \bibinfo{volume}{99} (\bibinfo{year}{2025}), \bibinfo{pages}{103334}.
\newblock


\bibitem[Jiang et~al\mbox{.}(2021)]%
        {jiang2021enlightengan}
\bibfield{author}{\bibinfo{person}{Yifan Jiang}, \bibinfo{person}{Xinyu Gong}, \bibinfo{person}{Ding Liu}, \bibinfo{person}{Yu Cheng}, \bibinfo{person}{Chen Fang}, \bibinfo{person}{Xiaohui Shen}, \bibinfo{person}{Jianchao Yang}, \bibinfo{person}{Pan Zhou}, {and} \bibinfo{person}{Zhangyang Wang}.} \bibinfo{year}{2021}\natexlab{}.
\newblock \showarticletitle{Enlightengan: Deep light enhancement without paired supervision}.
\newblock \bibinfo{journal}{\emph{IEEE transactions on image processing}}  \bibinfo{volume}{30} (\bibinfo{year}{2021}), \bibinfo{pages}{2340--2349}.
\newblock


\bibitem[Jin et~al\mbox{.}(2022)]%
        {jin2022unsupervised}
\bibfield{author}{\bibinfo{person}{Yeying Jin}, \bibinfo{person}{Wenhan Yang}, {and} \bibinfo{person}{Robby~T Tan}.} \bibinfo{year}{2022}\natexlab{}.
\newblock \showarticletitle{Unsupervised night image enhancement: When layer decomposition meets light-effects suppression}. In \bibinfo{booktitle}{\emph{European Conference on Computer Vision}}. Springer, \bibinfo{pages}{404--421}.
\newblock


\bibitem[Kingma and Ba(2014)]%
        {kingma2014adam}
\bibfield{author}{\bibinfo{person}{Diederik~P Kingma} {and} \bibinfo{person}{Jimmy Ba}.} \bibinfo{year}{2014}\natexlab{}.
\newblock \showarticletitle{Adam: A method for stochastic optimization}.
\newblock \bibinfo{journal}{\emph{arXiv preprint arXiv:1412.6980}} (\bibinfo{year}{2014}).
\newblock


\bibitem[Land and McCann(1971)]%
        {land1971lightness}
\bibfield{author}{\bibinfo{person}{Edwin~H Land} {and} \bibinfo{person}{John~J McCann}.} \bibinfo{year}{1971}\natexlab{}.
\newblock \showarticletitle{Lightness and retinex theory}.
\newblock \bibinfo{journal}{\emph{Journal of the Optical society of America}} \bibinfo{volume}{61}, \bibinfo{number}{1} (\bibinfo{year}{1971}), \bibinfo{pages}{1--11}.
\newblock


\bibitem[Li et~al\mbox{.}(2021)]%
        {li2021learning}
\bibfield{author}{\bibinfo{person}{Chongyi Li}, \bibinfo{person}{Chunle Guo}, {and} \bibinfo{person}{Chen~Change Loy}.} \bibinfo{year}{2021}\natexlab{}.
\newblock \showarticletitle{Learning to enhance low-light image via zero-reference deep curve estimation}.
\newblock \bibinfo{journal}{\emph{IEEE transactions on pattern analysis and machine intelligence}} \bibinfo{volume}{44}, \bibinfo{number}{8} (\bibinfo{year}{2021}), \bibinfo{pages}{4225--4238}.
\newblock


\bibitem[Li et~al\mbox{.}(2018)]%
        {li2018lightennet}
\bibfield{author}{\bibinfo{person}{Chongyi Li}, \bibinfo{person}{Jichang Guo}, \bibinfo{person}{Fatih Porikli}, {and} \bibinfo{person}{Yanwei Pang}.} \bibinfo{year}{2018}\natexlab{}.
\newblock \showarticletitle{LightenNet: A convolutional neural network for weakly illuminated image enhancement}.
\newblock \bibinfo{journal}{\emph{Pattern recognition letters}}  \bibinfo{volume}{104} (\bibinfo{year}{2018}), \bibinfo{pages}{15--22}.
\newblock


\bibitem[Li et~al\mbox{.}(2020)]%
        {li2020luminance}
\bibfield{author}{\bibinfo{person}{Jiaqian Li}, \bibinfo{person}{Juncheng Li}, \bibinfo{person}{Faming Fang}, \bibinfo{person}{Fang Li}, {and} \bibinfo{person}{Guixu Zhang}.} \bibinfo{year}{2020}\natexlab{}.
\newblock \showarticletitle{Luminance-aware pyramid network for low-light image enhancement}.
\newblock \bibinfo{journal}{\emph{IEEE Transactions on Multimedia}}  \bibinfo{volume}{23} (\bibinfo{year}{2020}), \bibinfo{pages}{3153--3165}.
\newblock


\bibitem[Li et~al\mbox{.}(2015)]%
        {liSoftBinarySegmentationbased2015}
\bibfield{author}{\bibinfo{person}{Zhenhao Li}, \bibinfo{person}{Kai Cheng}, {and} \bibinfo{person}{Xiaolin Wu}.} \bibinfo{year}{2015}\natexlab{}.
\newblock \showarticletitle{{Soft binary segmentation-based backlit image enhancement}}. In \bibinfo{booktitle}{\emph{{2015 IEEE 17th International Workshop on Multimedia Signal Processing (MMSP)}}}. \bibinfo{pages}{1--5}.
\newblock
\href{https://doi.org/10.1109/MMSP.2015.7340808}{doi:\nolinkurl{10.1109/MMSP.2015.7340808}}


\bibitem[Li and Wu(2017)]%
        {li2017learning}
\bibfield{author}{\bibinfo{person}{Zhenhao Li} {and} \bibinfo{person}{Xiaolin Wu}.} \bibinfo{year}{2017}\natexlab{}.
\newblock \showarticletitle{Learning-based restoration of backlit images}.
\newblock \bibinfo{journal}{\emph{IEEE Transactions on Image Processing}} \bibinfo{volume}{27}, \bibinfo{number}{2} (\bibinfo{year}{2017}), \bibinfo{pages}{976--986}.
\newblock


\bibitem[Liang et~al\mbox{.}(2022)]%
        {liang2022self}
\bibfield{author}{\bibinfo{person}{Jinxiu Liang}, \bibinfo{person}{Yong Xu}, \bibinfo{person}{Yuhui Quan}, \bibinfo{person}{Boxin Shi}, {and} \bibinfo{person}{Hui Ji}.} \bibinfo{year}{2022}\natexlab{}.
\newblock \showarticletitle{Self-supervised low-light image enhancement using discrepant untrained network priors}.
\newblock \bibinfo{journal}{\emph{IEEE Transactions on Circuits and Systems for Video Technology}} \bibinfo{volume}{32}, \bibinfo{number}{11} (\bibinfo{year}{2022}), \bibinfo{pages}{7332--7345}.
\newblock


\bibitem[Liang et~al\mbox{.}(2023)]%
        {liang2023iterative}
\bibfield{author}{\bibinfo{person}{Zhexin Liang}, \bibinfo{person}{Chongyi Li}, \bibinfo{person}{Shangchen Zhou}, \bibinfo{person}{Ruicheng Feng}, {and} \bibinfo{person}{Chen~Change Loy}.} \bibinfo{year}{2023}\natexlab{}.
\newblock \showarticletitle{Iterative prompt learning for unsupervised backlit image enhancement}. In \bibinfo{booktitle}{\emph{Proceedings of the IEEE/CVF International Conference on Computer Vision}}. \bibinfo{pages}{8094--8103}.
\newblock


\bibitem[Lim and Kim(2020)]%
        {lim2020dslr}
\bibfield{author}{\bibinfo{person}{Seokjae Lim} {and} \bibinfo{person}{Wonjun Kim}.} \bibinfo{year}{2020}\natexlab{}.
\newblock \showarticletitle{DSLR: Deep stacked Laplacian restorer for low-light image enhancement}.
\newblock \bibinfo{journal}{\emph{IEEE Transactions on Multimedia}}  \bibinfo{volume}{23} (\bibinfo{year}{2020}), \bibinfo{pages}{4272--4284}.
\newblock


\bibitem[Liu et~al\mbox{.}(2021)]%
        {liu2021retinex}
\bibfield{author}{\bibinfo{person}{Risheng Liu}, \bibinfo{person}{Long Ma}, \bibinfo{person}{Jiaao Zhang}, \bibinfo{person}{Xin Fan}, {and} \bibinfo{person}{Zhongxuan Luo}.} \bibinfo{year}{2021}\natexlab{}.
\newblock \showarticletitle{Retinex-inspired unrolling with cooperative prior architecture search for low-light image enhancement}. In \bibinfo{booktitle}{\emph{Proceedings of the IEEE/CVF conference on computer vision and pattern recognition}}. \bibinfo{pages}{10561--10570}.
\newblock


\bibitem[Liu et~al\mbox{.}(2024)]%
        {liu2024vmamba}
\bibfield{author}{\bibinfo{person}{Yue Liu}, \bibinfo{person}{Yunjie Tian}, \bibinfo{person}{Yuzhong Zhao}, \bibinfo{person}{Hongtian Yu}, \bibinfo{person}{Lingxi Xie}, \bibinfo{person}{Yaowei Wang}, \bibinfo{person}{Qixiang Ye}, \bibinfo{person}{Jianbin Jiao}, {and} \bibinfo{person}{Yunfan Liu}.} \bibinfo{year}{2024}\natexlab{}.
\newblock \showarticletitle{Vmamba: Visual state space model}.
\newblock \bibinfo{journal}{\emph{Advances in neural information processing systems}}  \bibinfo{volume}{37} (\bibinfo{year}{2024}), \bibinfo{pages}{103031--103063}.
\newblock


\bibitem[Lore et~al\mbox{.}(2017)]%
        {lore2017llnet}
\bibfield{author}{\bibinfo{person}{Kin~Gwn Lore}, \bibinfo{person}{Adedotun Akintayo}, {and} \bibinfo{person}{Soumik Sarkar}.} \bibinfo{year}{2017}\natexlab{}.
\newblock \showarticletitle{LLNet: A deep autoencoder approach to natural low-light image enhancement}.
\newblock \bibinfo{journal}{\emph{Pattern Recognition}}  \bibinfo{volume}{61} (\bibinfo{year}{2017}), \bibinfo{pages}{650--662}.
\newblock


\bibitem[Lu and Zhang(2020)]%
        {lu2020tbefn}
\bibfield{author}{\bibinfo{person}{Kun Lu} {and} \bibinfo{person}{Lihong Zhang}.} \bibinfo{year}{2020}\natexlab{}.
\newblock \showarticletitle{TBEFN: A two-branch exposure-fusion network for low-light image enhancement}.
\newblock \bibinfo{journal}{\emph{IEEE Transactions on Multimedia}}  \bibinfo{volume}{23} (\bibinfo{year}{2020}), \bibinfo{pages}{4093--4105}.
\newblock


\bibitem[Lv et~al\mbox{.}(2020)]%
        {lv2020fast}
\bibfield{author}{\bibinfo{person}{Feifan Lv}, \bibinfo{person}{Bo Liu}, {and} \bibinfo{person}{Feng Lu}.} \bibinfo{year}{2020}\natexlab{}.
\newblock \showarticletitle{Fast enhancement for non-uniform illumination images using light-weight CNNs}. In \bibinfo{booktitle}{\emph{Proceedings of the 28th ACM International Conference on Multimedia}}. \bibinfo{pages}{1450--1458}.
\newblock


\bibitem[Lv et~al\mbox{.}(2018)]%
        {lv2018mbllen}
\bibfield{author}{\bibinfo{person}{Feifan Lv}, \bibinfo{person}{Feng Lu}, \bibinfo{person}{Jianhua Wu}, {and} \bibinfo{person}{Chongsoon Lim}.} \bibinfo{year}{2018}\natexlab{}.
\newblock \showarticletitle{MBLLEN: Low-light image/video enhancement using cnns.}. In \bibinfo{booktitle}{\emph{Bmvc}}, Vol.~\bibinfo{volume}{220}. Northumbria University, \bibinfo{pages}{4}.
\newblock


\bibitem[Lv et~al\mbox{.}(2022)]%
        {lvBacklitNetDatasetNetwork2022}
\bibfield{author}{\bibinfo{person}{Xiaoqian Lv}, \bibinfo{person}{Shengping Zhang}, \bibinfo{person}{Qinglin Liu}, \bibinfo{person}{Haozhe Xie}, \bibinfo{person}{Bineng Zhong}, {and} \bibinfo{person}{Huiyu Zhou}.} \bibinfo{year}{2022}\natexlab{}.
\newblock \showarticletitle{{{BacklitNet}}: {{A}} Dataset and Network for Backlit Image Enhancement}.
\newblock \bibinfo{journal}{\emph{Computer Vision and Image Understanding}}  \bibinfo{volume}{218} (\bibinfo{date}{April} \bibinfo{year}{2022}), \bibinfo{pages}{103403}.
\newblock
\showISSN{1077-3142}
\href{https://doi.org/10.1016/j.cviu.2022.103403}{doi:\nolinkurl{10.1016/j.cviu.2022.103403}}


\bibitem[Ma et~al\mbox{.}(2022)]%
        {ma2022toward}
\bibfield{author}{\bibinfo{person}{Long Ma}, \bibinfo{person}{Tengyu Ma}, \bibinfo{person}{Risheng Liu}, \bibinfo{person}{Xin Fan}, {and} \bibinfo{person}{Zhongxuan Luo}.} \bibinfo{year}{2022}\natexlab{}.
\newblock \showarticletitle{Toward fast, flexible, and robust low-light image enhancement}. In \bibinfo{booktitle}{\emph{Proceedings of the IEEE/CVF conference on computer vision and pattern recognition}}. \bibinfo{pages}{5637--5646}.
\newblock


\bibitem[Paszke et~al\mbox{.}(2019)]%
        {10.5555/3454287.3455008}
\bibfield{author}{\bibinfo{person}{Adam Paszke}, \bibinfo{person}{Sam Gross}, \bibinfo{person}{Francisco Massa}, \bibinfo{person}{Adam Lerer}, \bibinfo{person}{James Bradbury}, \bibinfo{person}{Gregory Chanan}, \bibinfo{person}{Trevor Killeen}, \bibinfo{person}{Zeming Lin}, \bibinfo{person}{Natalia Gimelshein}, \bibinfo{person}{Luca Antiga}, \bibinfo{person}{Alban Desmaison}, \bibinfo{person}{Andreas K\"{o}pf}, \bibinfo{person}{Edward Yang}, \bibinfo{person}{Zach DeVito}, \bibinfo{person}{Martin Raison}, \bibinfo{person}{Alykhan Tejani}, \bibinfo{person}{Sasank Chilamkurthy}, \bibinfo{person}{Benoit Steiner}, \bibinfo{person}{Lu Fang}, \bibinfo{person}{Junjie Bai}, {and} \bibinfo{person}{Soumith Chintala}.} \bibinfo{year}{2019}\natexlab{}.
\newblock \bibinfo{booktitle}{\emph{PyTorch: an imperative style, high-performance deep learning library}}.
\newblock \bibinfo{publisher}{Curran Associates Inc.}, \bibinfo{address}{Red Hook, NY, USA}.
\newblock


\bibitem[Pei et~al\mbox{.}(2024)]%
        {pei2024efficientvmamba}
\bibfield{author}{\bibinfo{person}{Xiaohuan Pei}, \bibinfo{person}{Tao Huang}, {and} \bibinfo{person}{Chang Xu}.} \bibinfo{year}{2024}\natexlab{}.
\newblock \showarticletitle{Efficientvmamba: Atrous selective scan for light weight visual mamba}.
\newblock \bibinfo{journal}{\emph{arXiv preprint arXiv:2403.09977}} (\bibinfo{year}{2024}).
\newblock


\bibitem[Pisano et~al\mbox{.}(1998)]%
        {pisano1998contrast}
\bibfield{author}{\bibinfo{person}{Etta~D Pisano}, \bibinfo{person}{Shuquan Zong}, \bibinfo{person}{Bradley~M Hemminger}, \bibinfo{person}{Marla DeLuca}, \bibinfo{person}{R~Eugene Johnston}, \bibinfo{person}{Keith Muller}, \bibinfo{person}{M~Patricia Braeuning}, {and} \bibinfo{person}{Stephen~M Pizer}.} \bibinfo{year}{1998}\natexlab{}.
\newblock \showarticletitle{Contrast limited adaptive histogram equalization image processing to improve the detection of simulated spiculations in dense mammograms}.
\newblock \bibinfo{journal}{\emph{Journal of Digital imaging}}  \bibinfo{volume}{11} (\bibinfo{year}{1998}), \bibinfo{pages}{193--200}.
\newblock


\bibitem[Pizer et~al\mbox{.}(1987)]%
        {pizer1987adaptive}
\bibfield{author}{\bibinfo{person}{Stephen~M Pizer}, \bibinfo{person}{E~Philip Amburn}, \bibinfo{person}{John~D Austin}, \bibinfo{person}{Robert Cromartie}, \bibinfo{person}{Ari Geselowitz}, \bibinfo{person}{Trey Greer}, \bibinfo{person}{Bart ter Haar~Romeny}, \bibinfo{person}{John~B Zimmerman}, {and} \bibinfo{person}{Karel Zuiderveld}.} \bibinfo{year}{1987}\natexlab{}.
\newblock \showarticletitle{Adaptive histogram equalization and its variations}.
\newblock \bibinfo{journal}{\emph{Computer vision, graphics, and image processing}} \bibinfo{volume}{39}, \bibinfo{number}{3} (\bibinfo{year}{1987}), \bibinfo{pages}{355--368}.
\newblock


\bibitem[Ren et~al\mbox{.}(2019)]%
        {ren2019low}
\bibfield{author}{\bibinfo{person}{Wenqi Ren}, \bibinfo{person}{Sifei Liu}, \bibinfo{person}{Lin Ma}, \bibinfo{person}{Qianqian Xu}, \bibinfo{person}{Xiangyu Xu}, \bibinfo{person}{Xiaochun Cao}, \bibinfo{person}{Junping Du}, {and} \bibinfo{person}{Ming-Hsuan Yang}.} \bibinfo{year}{2019}\natexlab{}.
\newblock \showarticletitle{Low-light image enhancement via a deep hybrid network}.
\newblock \bibinfo{journal}{\emph{IEEE Transactions on Image Processing}} \bibinfo{volume}{28}, \bibinfo{number}{9} (\bibinfo{year}{2019}), \bibinfo{pages}{4364--4375}.
\newblock


\bibitem[Reza(2004)]%
        {reza2004realization}
\bibfield{author}{\bibinfo{person}{Ali~M Reza}.} \bibinfo{year}{2004}\natexlab{}.
\newblock \showarticletitle{Realization of the contrast limited adaptive histogram equalization (CLAHE) for real-time image enhancement}.
\newblock \bibinfo{journal}{\emph{Journal of VLSI signal processing systems for signal, image and video technology}}  \bibinfo{volume}{38} (\bibinfo{year}{2004}), \bibinfo{pages}{35--44}.
\newblock


\bibitem[Shazeer et~al\mbox{.}(2017)]%
        {shazeer2017outrageously}
\bibfield{author}{\bibinfo{person}{Noam Shazeer}, \bibinfo{person}{Azalia Mirhoseini}, \bibinfo{person}{Krzysztof Maziarz}, \bibinfo{person}{Andy Davis}, \bibinfo{person}{Quoc Le}, \bibinfo{person}{Geoffrey Hinton}, {and} \bibinfo{person}{Jeff Dean}.} \bibinfo{year}{2017}\natexlab{}.
\newblock \showarticletitle{Outrageously large neural networks: The sparsely-gated mixture-of-experts layer}.
\newblock \bibinfo{journal}{\emph{arXiv preprint arXiv:1701.06538}} (\bibinfo{year}{2017}).
\newblock


\bibitem[Smith(1978)]%
        {smith1978color}
\bibfield{author}{\bibinfo{person}{Alvy~Ray Smith}.} \bibinfo{year}{1978}\natexlab{}.
\newblock \showarticletitle{Color gamut transform pairs}.
\newblock \bibinfo{journal}{\emph{ACM Siggraph Computer Graphics}} \bibinfo{volume}{12}, \bibinfo{number}{3} (\bibinfo{year}{1978}), \bibinfo{pages}{12--19}.
\newblock


\bibitem[Triantafyllidou et~al\mbox{.}(2020)]%
        {triantafyllidou2020low}
\bibfield{author}{\bibinfo{person}{Danai Triantafyllidou}, \bibinfo{person}{Sean Moran}, \bibinfo{person}{Steven McDonagh}, \bibinfo{person}{Sarah Parisot}, {and} \bibinfo{person}{Gregory Slabaugh}.} \bibinfo{year}{2020}\natexlab{}.
\newblock \showarticletitle{Low light video enhancement using synthetic data produced with an intermediate domain mapping}. In \bibinfo{booktitle}{\emph{Computer vision--ECCV 2020: 16th European conference, glasgow, UK, August 23--28, 2020, proceedings, part XIII 16}}. Springer, \bibinfo{pages}{103--119}.
\newblock


\bibitem[Trongtirakul et~al\mbox{.}(2020)]%
        {trongtirakulSingleBacklitImage2020}
\bibfield{author}{\bibinfo{person}{Thaweesak Trongtirakul}, \bibinfo{person}{Werapon Chiracharit}, {and} \bibinfo{person}{Sos~S. Agaian}.} \bibinfo{year}{2020}\natexlab{}.
\newblock \showarticletitle{Single {{Backlit Image Enhancement}}}.
\newblock \bibinfo{journal}{\emph{IEEE Access}}  \bibinfo{volume}{8} (\bibinfo{year}{2020}), \bibinfo{pages}{71940--71950}.
\newblock
\showISSN{2169-3536}
\href{https://doi.org/10.1109/ACCESS.2020.2987256}{doi:\nolinkurl{10.1109/ACCESS.2020.2987256}}


\bibitem[Vaswani et~al\mbox{.}(2017)]%
        {vaswani2017attention}
\bibfield{author}{\bibinfo{person}{Ashish Vaswani}, \bibinfo{person}{Noam Shazeer}, \bibinfo{person}{Niki Parmar}, \bibinfo{person}{Jakob Uszkoreit}, \bibinfo{person}{Llion Jones}, \bibinfo{person}{Aidan~N Gomez}, \bibinfo{person}{{\L}ukasz Kaiser}, {and} \bibinfo{person}{Illia Polosukhin}.} \bibinfo{year}{2017}\natexlab{}.
\newblock \showarticletitle{Attention is all you need}.
\newblock \bibinfo{journal}{\emph{Advances in neural information processing systems}}  \bibinfo{volume}{30} (\bibinfo{year}{2017}).
\newblock


\bibitem[{Vazquez-Corral} et~al\mbox{.}(2018)]%
        {vazquez-corralPerceptuallybasedRestorationBacklit2018}
\bibfield{author}{\bibinfo{person}{Javier {Vazquez-Corral}}, \bibinfo{person}{Praveen Cyriac}, {and} \bibinfo{person}{Marcelo Bertalm{\'i}o}.} \bibinfo{year}{2018}\natexlab{}.
\newblock \showarticletitle{Perceptually-Based Restoration of Backlit Images}.
\newblock \bibinfo{journal}{\emph{Color and Imaging Conference}} \bibinfo{volume}{26}, \bibinfo{number}{1} (\bibinfo{date}{Nov.} \bibinfo{year}{2018}), \bibinfo{pages}{32--37}.
\newblock
\showISSN{2166-9635}
\href{https://doi.org/10.2352/ISSN.2169-2629.2018.26.32}{doi:\nolinkurl{10.2352/ISSN.2169-2629.2018.26.32}}


\bibitem[Wang et~al\mbox{.}(2016)]%
        {wangFusionbasedMethodSingle2016}
\bibfield{author}{\bibinfo{person}{Qiuhong Wang}, \bibinfo{person}{Xueyang Fu}, \bibinfo{person}{Xiao-Ping Zhang}, {and} \bibinfo{person}{Xinghao Ding}.} \bibinfo{year}{2016}\natexlab{}.
\newblock \showarticletitle{A Fusion-Based Method for Single Backlit Image Enhancement}. In \bibinfo{booktitle}{\emph{2016 {{IEEE International Conference}} on {{Image Processing}} ({{ICIP}})}}. \bibinfo{pages}{4077--4081}.
\newblock
\showISSN{2381-8549}
\href{https://doi.org/10.1109/ICIP.2016.7533126}{doi:\nolinkurl{10.1109/ICIP.2016.7533126}}


\bibitem[Wang et~al\mbox{.}(2019)]%
        {wang2019underexposed}
\bibfield{author}{\bibinfo{person}{Ruixing Wang}, \bibinfo{person}{Qing Zhang}, \bibinfo{person}{Chi-Wing Fu}, \bibinfo{person}{Xiaoyong Shen}, \bibinfo{person}{Wei-Shi Zheng}, {and} \bibinfo{person}{Jiaya Jia}.} \bibinfo{year}{2019}\natexlab{}.
\newblock \showarticletitle{Underexposed photo enhancement using deep illumination estimation}. In \bibinfo{booktitle}{\emph{Proceedings of the IEEE/CVF conference on computer vision and pattern recognition}}. \bibinfo{pages}{6849--6857}.
\newblock


\bibitem[Wang et~al\mbox{.}(2024)]%
        {wang2024dpec}
\bibfield{author}{\bibinfo{person}{Shuang Wang}, \bibinfo{person}{Qianwen Lu}, \bibinfo{person}{Boxing Peng}, \bibinfo{person}{Yihe Nie}, {and} \bibinfo{person}{Qingchuan Tao}.} \bibinfo{year}{2024}\natexlab{}.
\newblock \showarticletitle{DPEC: Dual-Path Error Compensation Method for Enhanced Low-Light Image Clarity}.
\newblock \bibinfo{journal}{\emph{arXiv preprint arXiv:2407.09553}} (\bibinfo{year}{2024}).
\newblock


\bibitem[Wang et~al\mbox{.}(2023)]%
        {wang2023ultra}
\bibfield{author}{\bibinfo{person}{Tao Wang}, \bibinfo{person}{Kaihao Zhang}, \bibinfo{person}{Tianrun Shen}, \bibinfo{person}{Wenhan Luo}, \bibinfo{person}{Bjorn Stenger}, {and} \bibinfo{person}{Tong Lu}.} \bibinfo{year}{2023}\natexlab{}.
\newblock \showarticletitle{Ultra-high-definition low-light image enhancement: A benchmark and transformer-based method}. In \bibinfo{booktitle}{\emph{Proceedings of the AAAI conference on artificial intelligence}}, Vol.~\bibinfo{volume}{37}. \bibinfo{pages}{2654--2662}.
\newblock


\bibitem[Wang et~al\mbox{.}(2022)]%
        {wang2022low}
\bibfield{author}{\bibinfo{person}{Yufei Wang}, \bibinfo{person}{Renjie Wan}, \bibinfo{person}{Wenhan Yang}, \bibinfo{person}{Haoliang Li}, \bibinfo{person}{Lap-Pui Chau}, {and} \bibinfo{person}{Alex Kot}.} \bibinfo{year}{2022}\natexlab{}.
\newblock \showarticletitle{Low-light image enhancement with normalizing flow}. In \bibinfo{booktitle}{\emph{Proceedings of the AAAI conference on artificial intelligence}}, Vol.~\bibinfo{volume}{36}. \bibinfo{pages}{2604--2612}.
\newblock


\bibitem[Wang et~al\mbox{.}(2004)]%
        {wang2004image}
\bibfield{author}{\bibinfo{person}{Zhou Wang}, \bibinfo{person}{Alan~C Bovik}, \bibinfo{person}{Hamid~R Sheikh}, {and} \bibinfo{person}{Eero~P Simoncelli}.} \bibinfo{year}{2004}\natexlab{}.
\newblock \showarticletitle{Image quality assessment: from error visibility to structural similarity}.
\newblock \bibinfo{journal}{\emph{IEEE transactions on image processing}} \bibinfo{volume}{13}, \bibinfo{number}{4} (\bibinfo{year}{2004}), \bibinfo{pages}{600--612}.
\newblock


\bibitem[Wei et~al\mbox{.}(2018)]%
        {wei2018deep}
\bibfield{author}{\bibinfo{person}{Chen Wei}, \bibinfo{person}{Wenjing Wang}, \bibinfo{person}{Wenhan Yang}, {and} \bibinfo{person}{Jiaying Liu}.} \bibinfo{year}{2018}\natexlab{}.
\newblock \showarticletitle{Deep retinex decomposition for low-light enhancement}.
\newblock \bibinfo{journal}{\emph{arXiv preprint arXiv:1808.04560}} (\bibinfo{year}{2018}).
\newblock


\bibitem[Weng et~al\mbox{.}(2024)]%
        {weng2024mamballie}
\bibfield{author}{\bibinfo{person}{Jiangwei Weng}, \bibinfo{person}{Zhiqiang Yan}, \bibinfo{person}{Ying Tai}, \bibinfo{person}{Jianjun Qian}, \bibinfo{person}{Jian Yang}, {and} \bibinfo{person}{Jun Li}.} \bibinfo{year}{2024}\natexlab{}.
\newblock \showarticletitle{Mamballie: Implicit retinex-aware low light enhancement with global-then-local state space}.
\newblock \bibinfo{journal}{\emph{arXiv preprint arXiv:2405.16105}} (\bibinfo{year}{2024}).
\newblock


\bibitem[Wu et~al\mbox{.}(2021)]%
        {wu2021visual}
\bibfield{author}{\bibinfo{person}{Bichen Wu}, \bibinfo{person}{Chenfeng Xu}, \bibinfo{person}{Xiaoliang Dai}, \bibinfo{person}{Alvin Wan}, \bibinfo{person}{Peizhao Zhang}, \bibinfo{person}{Zhicheng Yan}, \bibinfo{person}{Masayoshi Tomizuka}, \bibinfo{person}{Joseph~E Gonzalez}, \bibinfo{person}{Kurt Keutzer}, {and} \bibinfo{person}{Peter Vajda}.} \bibinfo{year}{2021}\natexlab{}.
\newblock \showarticletitle{Visual transformers: Where do transformers really belong in vision models?}. In \bibinfo{booktitle}{\emph{Proceedings of the IEEE/CVF International Conference on Computer Vision}}. \bibinfo{pages}{599--609}.
\newblock


\bibitem[Wu et~al\mbox{.}(2022)]%
        {wu2022uretinex}
\bibfield{author}{\bibinfo{person}{Wenhui Wu}, \bibinfo{person}{Jian Weng}, \bibinfo{person}{Pingping Zhang}, \bibinfo{person}{Xu Wang}, \bibinfo{person}{Wenhan Yang}, {and} \bibinfo{person}{Jianmin Jiang}.} \bibinfo{year}{2022}\natexlab{}.
\newblock \showarticletitle{Uretinex-net: Retinex-based deep unfolding network for low-light image enhancement}. In \bibinfo{booktitle}{\emph{Proceedings of the IEEE/CVF conference on computer vision and pattern recognition}}. \bibinfo{pages}{5901--5910}.
\newblock


\bibitem[Xiao et~al\mbox{.}(2019)]%
        {xiao2019histogram}
\bibfield{author}{\bibinfo{person}{Bin Xiao}, \bibinfo{person}{Yunqiu Xu}, \bibinfo{person}{Han Tang}, \bibinfo{person}{Xiuli Bi}, {and} \bibinfo{person}{Weisheng Li}.} \bibinfo{year}{2019}\natexlab{}.
\newblock \showarticletitle{Histogram learning in image contrast enhancement}. In \bibinfo{booktitle}{\emph{Proceedings of the IEEE/CVF Conference on Computer Vision and Pattern Recognition Workshops}}. \bibinfo{pages}{0--0}.
\newblock


\bibitem[Xu et~al\mbox{.}(2020)]%
        {xu2020learning}
\bibfield{author}{\bibinfo{person}{Ke Xu}, \bibinfo{person}{Xin Yang}, \bibinfo{person}{Baocai Yin}, {and} \bibinfo{person}{Rynson~WH Lau}.} \bibinfo{year}{2020}\natexlab{}.
\newblock \showarticletitle{Learning to restore low-light images via decomposition-and-enhancement}. In \bibinfo{booktitle}{\emph{Proceedings of the IEEE/CVF conference on computer vision and pattern recognition}}. \bibinfo{pages}{2281--2290}.
\newblock


\bibitem[Xu et~al\mbox{.}(2024)]%
        {xu2024visual}
\bibfield{author}{\bibinfo{person}{Rui Xu}, \bibinfo{person}{Shu Yang}, \bibinfo{person}{Yihui Wang}, \bibinfo{person}{Yu Cai}, \bibinfo{person}{Bo Du}, {and} \bibinfo{person}{Hao Chen}.} \bibinfo{year}{2024}\natexlab{}.
\newblock \showarticletitle{Visual mamba: A survey and new outlooks}.
\newblock \bibinfo{journal}{\emph{arXiv preprint arXiv:2404.18861}} (\bibinfo{year}{2024}).
\newblock


\bibitem[Xu et~al\mbox{.}(2022)]%
        {xu2022snr}
\bibfield{author}{\bibinfo{person}{Xiaogang Xu}, \bibinfo{person}{Ruixing Wang}, \bibinfo{person}{Chi-Wing Fu}, {and} \bibinfo{person}{Jiaya Jia}.} \bibinfo{year}{2022}\natexlab{}.
\newblock \showarticletitle{Snr-aware low-light image enhancement}. In \bibinfo{booktitle}{\emph{Proceedings of the IEEE/CVF conference on computer vision and pattern recognition}}. \bibinfo{pages}{17714--17724}.
\newblock


\bibitem[Yang et~al\mbox{.}(2023)]%
        {yang2023learning}
\bibfield{author}{\bibinfo{person}{Kai-Fu Yang}, \bibinfo{person}{Cheng Cheng}, \bibinfo{person}{Shi-Xuan Zhao}, \bibinfo{person}{Hong-Mei Yan}, \bibinfo{person}{Xian-Shi Zhang}, {and} \bibinfo{person}{Yong-Jie Li}.} \bibinfo{year}{2023}\natexlab{}.
\newblock \showarticletitle{Learning to adapt to light}.
\newblock \bibinfo{journal}{\emph{International Journal of Computer Vision}} \bibinfo{volume}{131}, \bibinfo{number}{4} (\bibinfo{year}{2023}), \bibinfo{pages}{1022--1041}.
\newblock


\bibitem[Yang et~al\mbox{.}(2020)]%
        {yang2020fidelity}
\bibfield{author}{\bibinfo{person}{Wenhan Yang}, \bibinfo{person}{Shiqi Wang}, \bibinfo{person}{Yuming Fang}, \bibinfo{person}{Yue Wang}, {and} \bibinfo{person}{Jiaying Liu}.} \bibinfo{year}{2020}\natexlab{}.
\newblock \showarticletitle{From fidelity to perceptual quality: A semi-supervised approach for low-light image enhancement}. In \bibinfo{booktitle}{\emph{Proceedings of the IEEE/CVF conference on computer vision and pattern recognition}}. \bibinfo{pages}{3063--3072}.
\newblock


\bibitem[Ye et~al\mbox{.}(2023)]%
        {ye2023spatio}
\bibfield{author}{\bibinfo{person}{Jing Ye}, \bibinfo{person}{Changzhen Qiu}, {and} \bibinfo{person}{Zhiyong Zhang}.} \bibinfo{year}{2023}\natexlab{}.
\newblock \showarticletitle{Spatio-temporal propagation and reconstruction for low-light video enhancement}.
\newblock \bibinfo{journal}{\emph{Digital Signal Processing}}  \bibinfo{volume}{139} (\bibinfo{year}{2023}), \bibinfo{pages}{104071}.
\newblock


\bibitem[Ye et~al\mbox{.}(2024)]%
        {ye2024survey}
\bibfield{author}{\bibinfo{person}{Jing Ye}, \bibinfo{person}{Changzhen Qiu}, {and} \bibinfo{person}{Zhiyong Zhang}.} \bibinfo{year}{2024}\natexlab{}.
\newblock \showarticletitle{A survey on learning-based low-light image and video enhancement}.
\newblock \bibinfo{journal}{\emph{Displays}}  \bibinfo{volume}{81} (\bibinfo{year}{2024}), \bibinfo{pages}{102614}.
\newblock


\bibitem[Zamir et~al\mbox{.}(2022)]%
        {zamir2022restormer}
\bibfield{author}{\bibinfo{person}{Syed~Waqas Zamir}, \bibinfo{person}{Aditya Arora}, \bibinfo{person}{Salman Khan}, \bibinfo{person}{Munawar Hayat}, \bibinfo{person}{Fahad~Shahbaz Khan}, {and} \bibinfo{person}{Ming-Hsuan Yang}.} \bibinfo{year}{2022}\natexlab{}.
\newblock \showarticletitle{Restormer: Efficient transformer for high-resolution image restoration}. In \bibinfo{booktitle}{\emph{Proceedings of the IEEE/CVF conference on computer vision and pattern recognition}}. \bibinfo{pages}{5728--5739}.
\newblock


\bibitem[Zamir et~al\mbox{.}(2020)]%
        {zamirLearningEnrichedFeatures2020}
\bibfield{author}{\bibinfo{person}{Syed~Waqas Zamir}, \bibinfo{person}{Aditya Arora}, \bibinfo{person}{Salman Khan}, \bibinfo{person}{Munawar Hayat}, \bibinfo{person}{Fahad~Shahbaz Khan}, \bibinfo{person}{Ming-Hsuan Yang}, {and} \bibinfo{person}{Ling Shao}.} \bibinfo{year}{2020}\natexlab{}.
\newblock \showarticletitle{Learning {{Enriched Features}} for {{Real Image Restoration}} and {{Enhancement}}}. In \bibinfo{booktitle}{\emph{Computer {{Vision}} -- {{ECCV}} 2020}}, \bibfield{editor}{\bibinfo{person}{Andrea Vedaldi}, \bibinfo{person}{Horst Bischof}, \bibinfo{person}{Thomas Brox}, {and} \bibinfo{person}{Jan-Michael Frahm}} (Eds.). \bibinfo{publisher}{Springer International Publishing}, \bibinfo{address}{Cham}, \bibinfo{pages}{492--511}.
\newblock
\showISBNx{978-3-030-58595-2}
\href{https://doi.org/10.1007/978-3-030-58595-2_30}{doi:\nolinkurl{10.1007/978-3-030-58595-2_30}}


\bibitem[Zhang et~al\mbox{.}(2019b)]%
        {zhang2019zero}
\bibfield{author}{\bibinfo{person}{Lin Zhang}, \bibinfo{person}{Lijun Zhang}, \bibinfo{person}{Xiao Liu}, \bibinfo{person}{Ying Shen}, \bibinfo{person}{Shaoming Zhang}, {and} \bibinfo{person}{Shengjie Zhao}.} \bibinfo{year}{2019}\natexlab{b}.
\newblock \showarticletitle{Zero-shot restoration of back-lit images using deep internal learning}. In \bibinfo{booktitle}{\emph{Proceedings of the 27th ACM international conference on multimedia}}. \bibinfo{pages}{1623--1631}.
\newblock


\bibitem[Zhang et~al\mbox{.}(2018)]%
        {zhang2018unreasonable}
\bibfield{author}{\bibinfo{person}{Richard Zhang}, \bibinfo{person}{Phillip Isola}, \bibinfo{person}{Alexei~A Efros}, \bibinfo{person}{Eli Shechtman}, {and} \bibinfo{person}{Oliver Wang}.} \bibinfo{year}{2018}\natexlab{}.
\newblock \showarticletitle{The unreasonable effectiveness of deep features as a perceptual metric}. In \bibinfo{booktitle}{\emph{Proceedings of the IEEE conference on computer vision and pattern recognition}}. \bibinfo{pages}{586--595}.
\newblock


\bibitem[Zhang et~al\mbox{.}(2024)]%
        {zhang2024llemamba}
\bibfield{author}{\bibinfo{person}{Xuanqi Zhang}, \bibinfo{person}{Haijin Zeng}, \bibinfo{person}{Jinwang Pan}, \bibinfo{person}{Qiangqiang Shen}, {and} \bibinfo{person}{Yongyong Chen}.} \bibinfo{year}{2024}\natexlab{}.
\newblock \showarticletitle{LLEMamba: Low-Light Enhancement via Relighting-Guided Mamba with Deep Unfolding Network}.
\newblock \bibinfo{journal}{\emph{arXiv preprint arXiv:2406.01028}} (\bibinfo{year}{2024}).
\newblock


\bibitem[Zhang et~al\mbox{.}(2021)]%
        {zhang2021beyond}
\bibfield{author}{\bibinfo{person}{Yonghua Zhang}, \bibinfo{person}{Xiaojie Guo}, \bibinfo{person}{Jiayi Ma}, \bibinfo{person}{Wei Liu}, {and} \bibinfo{person}{Jiawan Zhang}.} \bibinfo{year}{2021}\natexlab{}.
\newblock \showarticletitle{Beyond brightening low-light images}.
\newblock \bibinfo{journal}{\emph{International Journal of Computer Vision}}  \bibinfo{volume}{129} (\bibinfo{year}{2021}), \bibinfo{pages}{1013--1037}.
\newblock


\bibitem[Zhang et~al\mbox{.}(2019a)]%
        {zhang2019kindling}
\bibfield{author}{\bibinfo{person}{Yonghua Zhang}, \bibinfo{person}{Jiawan Zhang}, {and} \bibinfo{person}{Xiaojie Guo}.} \bibinfo{year}{2019}\natexlab{a}.
\newblock \showarticletitle{Kindling the darkness: A practical low-light image enhancer}. In \bibinfo{booktitle}{\emph{Proceedings of the 27th ACM international conference on multimedia}}. \bibinfo{pages}{1632--1640}.
\newblock


\bibitem[Zhang et~al\mbox{.}(2022)]%
        {zhang2022deep}
\bibfield{author}{\bibinfo{person}{Zhao Zhang}, \bibinfo{person}{Huan Zheng}, \bibinfo{person}{Richang Hong}, \bibinfo{person}{Mingliang Xu}, \bibinfo{person}{Shuicheng Yan}, {and} \bibinfo{person}{Meng Wang}.} \bibinfo{year}{2022}\natexlab{}.
\newblock \showarticletitle{Deep color consistent network for low-light image enhancement}. In \bibinfo{booktitle}{\emph{Proceedings of the IEEE/CVF conference on computer vision and pattern recognition}}. \bibinfo{pages}{1899--1908}.
\newblock


\bibitem[Zhao et~al\mbox{.}(2021)]%
        {zhao2021retinexdip}
\bibfield{author}{\bibinfo{person}{Zunjin Zhao}, \bibinfo{person}{Bangshu Xiong}, \bibinfo{person}{Lei Wang}, \bibinfo{person}{Qiaofeng Ou}, \bibinfo{person}{Lei Yu}, {and} \bibinfo{person}{Fa Kuang}.} \bibinfo{year}{2021}\natexlab{}.
\newblock \showarticletitle{RetinexDIP: A unified deep framework for low-light image enhancement}.
\newblock \bibinfo{journal}{\emph{IEEE Transactions on Circuits and Systems for Video Technology}} \bibinfo{volume}{32}, \bibinfo{number}{3} (\bibinfo{year}{2021}), \bibinfo{pages}{1076--1088}.
\newblock


\bibitem[Zheng et~al\mbox{.}(2022)]%
        {zheng2022low}
\bibfield{author}{\bibinfo{person}{Shen Zheng}, \bibinfo{person}{Yiling Ma}, \bibinfo{person}{Jinqian Pan}, \bibinfo{person}{Changjie Lu}, {and} \bibinfo{person}{Gaurav Gupta}.} \bibinfo{year}{2022}\natexlab{}.
\newblock \showarticletitle{Low-light image and video enhancement: A comprehensive survey and beyond}.
\newblock \bibinfo{journal}{\emph{arXiv preprint arXiv:2212.10772}} (\bibinfo{year}{2022}).
\newblock


\bibitem[Zhu et~al\mbox{.}(2022)]%
        {zhu2022uni}
\bibfield{author}{\bibinfo{person}{Jinguo Zhu}, \bibinfo{person}{Xizhou Zhu}, \bibinfo{person}{Wenhai Wang}, \bibinfo{person}{Xiaohua Wang}, \bibinfo{person}{Hongsheng Li}, \bibinfo{person}{Xiaogang Wang}, {and} \bibinfo{person}{Jifeng Dai}.} \bibinfo{year}{2022}\natexlab{}.
\newblock \showarticletitle{Uni-perceiver-moe: Learning sparse generalist models with conditional moes}.
\newblock \bibinfo{journal}{\emph{Advances in Neural Information Processing Systems}}  \bibinfo{volume}{35} (\bibinfo{year}{2022}), \bibinfo{pages}{2664--2678}.
\newblock


\bibitem[Zhu et~al\mbox{.}(2020)]%
        {zhu2020eemefn}
\bibfield{author}{\bibinfo{person}{Minfeng Zhu}, \bibinfo{person}{Pingbo Pan}, \bibinfo{person}{Wei Chen}, {and} \bibinfo{person}{Yi Yang}.} \bibinfo{year}{2020}\natexlab{}.
\newblock \showarticletitle{Eemefn: Low-light image enhancement via edge-enhanced multi-exposure fusion network}. In \bibinfo{booktitle}{\emph{Proceedings of the AAAI conference on artificial intelligence}}, Vol.~\bibinfo{volume}{34}. \bibinfo{pages}{13106--13113}.
\newblock


\bibitem[Zou et~al\mbox{.}(2024a)]%
        {zouWaveMambaWaveletState2024}
\bibfield{author}{\bibinfo{person}{Wenbin Zou}, \bibinfo{person}{Hongxia Gao}, \bibinfo{person}{Weipeng Yang}, {and} \bibinfo{person}{Tongtong Liu}.} \bibinfo{year}{2024}\natexlab{a}.
\newblock \showarticletitle{Wave-{{Mamba}}: {{Wavelet State Space Model}} for {{Ultra-High-Definition Low-Light Image Enhancement}}}. In \bibinfo{booktitle}{\emph{Proceedings of the 32nd {{ACM International Conference}} on {{Multimedia}}}} \emph{(\bibinfo{series}{{{MM}} '24})}. \bibinfo{publisher}{Association for Computing Machinery}, \bibinfo{address}{New York, NY, USA}, \bibinfo{pages}{1534--1543}.
\newblock
\showISBNx{979-8-4007-0686-8}
\href{https://doi.org/10.1145/3664647.3681580}{doi:\nolinkurl{10.1145/3664647.3681580}}


\bibitem[Zou et~al\mbox{.}(2024b)]%
        {zou2024wave}
\bibfield{author}{\bibinfo{person}{Wenbin Zou}, \bibinfo{person}{Hongxia Gao}, \bibinfo{person}{Weipeng Yang}, {and} \bibinfo{person}{Tongtong Liu}.} \bibinfo{year}{2024}\natexlab{b}.
\newblock \showarticletitle{Wave-mamba: Wavelet state space model for ultra-high-definition low-light image enhancement}. In \bibinfo{booktitle}{\emph{Proceedings of the 32nd ACM International Conference on Multimedia}}. \bibinfo{pages}{1534--1543}.
\newblock


\end{thebibliography}










\end{document}